\RequirePackage{fix-cm}
\RequirePackage{amsmath}

\documentclass[twocolumn]{svjour3}
\smartqed  
\usepackage{url}
\usepackage{graphicx}
\usepackage{latexsym}
\usepackage{amssymb}
\usepackage[american]{babel}
\usepackage[footnote, nolist]{acronym}

\begin{document}

\acrodef{3D}{Three-Dimensional}
\acrodef{PL}{Piecewise Linear}
\acrodef{PS}{Piecewise Smooth}
\acrodef{SfM}{Structure from Motion}
\acrodef{KLT}{Kanade-Lucas-Tomasi}
\acrodef{VO}{Visual Odometry}
\acrodef{SVD}{Singular Value Decomposition}
\acrodef{RANSAC}{Random Sample Consensus}

\title{Complete Endomorphisms in Computer Vision}
\subtitle{Applied to multiple view geometry}

\author{Finat, Javier \and Delgado-del-Hoyo, Francisco}


\institute{MoBiVAP research group \\
          Universidad de Valladolid \at
          Campus Miguel Delibes \\
          Paseo de Bel\'en, 11 \\
          47010, Valladolid, Spain \\
          Tel.: +34-983-184-398 \\
          \url{https://www.mobivap.es} \\
          \email{franciscojavier.finat@uva.es}
}

\date{\today}

\maketitle

\begin{abstract}
Correspondences between $k$-tuples of points are key in multiple view geometry and motion analysis. Regular transformations are posed by homographies between two projective planes that serves as structural models for images. Such transformations can not include degenerate situations. Fundamental or essential matrices expand homographies with structural information by using degenerate bilinear maps. The projectivization of the endomorphisms of a three-dimensional vector space includes all of them. Hence, they are able to explain a wider range of eventually degenerate transformations between arbitrary pairs of views. To include these degenerate situations, this paper introduces a completion of bilinear maps between spaces given by an equivariant compactification of regular transformations. This completion is extensible to the varieties of fundamental and essential matrices, where most methods based on regular transformations fail. The construction of complete endomorphisms manages degenerate projection maps using a simultaneous action on source and target spaces. In such way, this mathematical construction provides a robust framework to relate corresponding views in multiple view geometry.
\keywords{Epipolar Geometry \and Essential Matrix \and Fundamental Matrix \and Degeneracies \and Secant Varieties \and Adjoint Representation}
\end{abstract}

\section{Introduction}
\label{sec:introduction}

A classical issue regarding 3D reconstruction and motion analysis concerns the preservation of the continuity of the scene or the flow, although small changes in input occurs. There are a lot of answers where different constraints have been introduced from the early eighties. The most common constraint is the structural bilinear - or multilinear - tensor created with $k$-tuples of corresponding elements (points and/or lines) for the camera pose. All these constraints are very sensitive to noise. Classical approaches are based on minimal solutions extracted from noise measurements following a \ac{RANSAC} scheme. However, indeterminacies persist for some degenerate situations - created by low rank matrices - that can arise for mobile cameras.

Classic literature in computer vision \cite{faugeras1993three,hartley2003multiple} display a low attention to degenerate cases in structural models. These cases arise when independence conditions for features are not fulfilled, including situations where the camera turns around its optical axis or it is in fronto-parallel position w.r.t. a planar surface (a wall or the ground, e.g.). Then, the problem is ill-posed, and conventional solutions consists in performing a ``small perturbation'' or reboot the process. Both strategies display issues concerning the lack of control about the perturbation to be made that generate undesirable discontinuities. Thus, it is important to develop alternative strategies which can maintain some kind of ``coherence'' by reusing the ``recent history'' of the trajectory. History is continuously modeled in terms of generically regular conditions for tensors in previously sampled images with a discrete approach of a well-defined path in the space of structural tensors. Unfortunately, degeneracy conditions for typical features give indeterminacy for limits of structural tensors, which must be removed. Our approach consists of considering Kinematic information of the matrix version of the gradient field for indeterminacy loci.

Less attention has been paid to preserve the ``continuity'' of eventually singular trajectories in the space of bilinear maps linked to the automatic correspondence between pairs of views. In this case, singular maps are the responsible for indeterminacies in tensors and lie on singular strata of the space of bilinear maps\cite{thorup1988complete}. In this work, we develop a more down-to-earth approach using some basic properties of the projectivization of spaces of endomorphisms, including homographies $\mathbf{H}$, fundamental $\mathbf{F}$ and essential $\mathbf{E}$ matrices. All of them can be described in terms of orbits by a group action on the space of endomorphisms $End(V)$, i.e. linear maps of a vector space $V$ in itself. Their simultaneous algebraic treatment allows to extend the algebraic completion to the space of eventually degenerated central projection maps $\mathbf{M}_{\mathbf{C}}$ with center $\mathbf{C}$. Intuitively, the key for the control of degenerate cases is to select appropriate limits of tangents in a ``more complete'' space.

Therefore, the \emph{main goal} of our work is to model a continuous solution that also considers degenerated cases for simplest tensors (fundamental and essential matrices, e.g.) such those appearing in structural models for 3D Reconstruction. In order to achieve this goal, we introduced an ``equivariant compactification'' of the space of matrices w.r.t. group actions linked to kinematic properties visualized in the dual space. This double representation (positions and ``speed'') stores the ``recent history'' represented by a path in the tangent bundle $\tau_{End(V)}$ to $End(V)$ for the trajectory of a mobile camera $\mathbf{C}(t)$. Our approach is based on a dual presentation for the rank stratification of matrices. This dual representation encodes tangential information at each point represented by the adjoint matrix, a $3 \times 3$-matrix whose entries represent the gradient $\nabla(det(\mathbf{A}))$ of the determinant of $\mathbf{A}$.

In the simplest case, after fixing a basis $B_V$ of $V$, endomorphisms of a 3D vector space $V \simeq \mathbb{R}^{3}$ are given by arbitrary $3 \times 3$-matrices; they are naturally stratified by the rank giving three orbits with rank $r\leq 3$. In particular, from the differential viewpoint, sets of homographies $\mathbf{H}$ and regular (i.e. rank 2) fundamental matrices $\mathbf{F}$ can be considered as two $G$-orbits of the Lie algebra $\mathfrak{g} = End(V^{3})$ of $G = GL(3) = Aut(V^{3})$ by the action of the projectivization $\mathbb{P}GL(3)$ of the general linear group corresponding to rank $3$ and rank $2$ matrices, respectively. More generally, the description of $End(V^{n+1})$ as a union of orbits by the action of $GL(n+1; \mathbb{R})$ gives a structure as an ``orbifold'', i.e. a union of $G$-orbits containing their degenerate cases, which are usually excluded from the analysis. They are recovered by introducing a ``compactification'' where degenerate cases are managed in terms of successive envelopes by linear subspaces $W$ of $V$. All arguments can be extended to higher dimensions and even to hypermatrices representing more sophisticated tensors. However, for simplification purposes, we constraint ourselves only to endomorphisms extending planar or spatial homographies to the singular case. 

The rest of the paper is organized as follows. Section \ref{sec:background} provides the mathematical background to understand the rest of the paper and frames our approach in the state of the art. Section \ref{sec:planar-homographies} analyzes the simplest cases including regular transformations defined by homographies for the planar case that relate 2D views using the fundamental variety. Section \ref{sec:algebraic-approach} extends the approach to rigid transformations in the third dimension, including metric aspects in terms of the essential variety involving source and target spaces in $\mathbb{P}^3$. Section \ref{sec:spatial-homographies} studies the structural connection between them using the simultaneous action on source and target spaces of a variable projection linked to the camera pose; left-right $\mathcal{A}$ and contact $\mathcal{K}$-equivalences are explained. Section \ref{sec:practical-considerations} provides additional insight concerning the details and practical considerations for implementing this approach in \ac{VO} systems. Finally, Section \ref{sec:conclusions} concludes the paper with a summary of the main results and guidelines for further research.

\section{Background}
\label{sec:background}

Local symmetries are ubiquitous in a lot of problems in Physics and Engineering involving propagation phenomena. Most approaches in applied areas consider only regular regions, by ignoring any kind of degenerations linked to rank deficient matrices linked to linearization of phenomena. To include them, we    develop a ``locally symmetric completion'' of eventually singular transformations for involved tensors such those appearing in Reconstruction issues. Our approach is not quite original; a similar idea can be found in \cite{tron2017space}, which introduces a locally symmetric structure in a differential framework concerning geodesics on the essential manifold. Nevertheless, the initial geometric description as symmetric space (union of orbits linked to the rank preservation) can not be extended to include a differential approach to degenerate cases. Due to the occurrence of singularities, the support given as a quotient variety is not a smooth manifold but a singular algebraic variety where the methods for Riemannian manifolds no longer apply.

A larger description of a locally symmetric structure may be performed by extending the ordinary algebraic approach. Roughly speaking, it suffices to add limits of tangent subspaces along different ``branches'' at singularities and ``extend the action'' to obtain a more complete description including kinematic aspects. In this way, one obtains a  ``local replication'' compatible with the presence of singularities in the adherence of ``augmented'' orbits by tangent spaces at regular points. So, for the ``subregular'' case - codimension one orbit - it suffices to construct pairs of eventually degenerate transformations involving the original one and a ``generalized dual'' transformation (given by the adjoint map in the regular case) representing neighbor tangent directions. So, first order differential approaches of eventually degenerate maps allow to propagate - and consequently, anticipate - partial representations of expected views, even in presence of rank-deficient matrices.

This extended duality allows a simultaneous treatment of incidence and tangency conditions (both are projectively invariant), and to manage degenerate cases in terms of ``complete objects'' as limits in an enlarged space (including the original space and their duals) which can be managed as a locally symmetric space in terms of extended transformations (original ones and their exterior powers). Besides its differential description as a gradient in the space of matrices, a more geometric description can be developed in terms of pairs of loci and their envelopes. The extended transformations act on the source or ambient space (right action), and on its dual space which can be considered as a target space representing envelopes by tangent subspaces. This idea is reminiscent of the contact action $\mathcal{K}$ which preserves the graph and it provides a natural extension of the right-left action $\mathcal{A}$ (see next paragraph). A discrete version of last action has been used by Kanade, Tomasi and Lucas \ac{KLT} along the early nineties in regard to Structure from Motion approaches to 3D Reconstruction. Both actions are commonly used for the infinitesimal classification of map-germs in differential classification of map-germs. However, our approach is more focused towards a local description of the space of generalized transformations and/or projection maps as a locally symmetric space. This structure has the additional advantage of allowing the extension of Riemannian properties given in terms of geodesics.

The simplest simultaneous action on source and target spaces is the Cartesian or direct product of actions. It is denoted by the $\mathcal{A}$-action where $\mathcal{A} := \mathcal{R} \times \mathcal{L}$ is the right-left action. Its orbits are given by the double conjugacy classes from the algebraic viewpoint. The $\mathcal{A}$-action is very useful for \emph{decoupled models} (implicit in \ac{KLT} algorithms or \ac{SfM}, e.g.), and consequently very useful by computational reasons. Despite the wide interest for the above approaches, the $\mathcal{A}$-action is less plausible than the $\mathcal{K}$-action, which incorporates the graph preservation (corresponding to quadratic contact between a manifold $M$ and its tangent space $T_p M$ at each contact point $p\in M$) as the structural constraint.

In our case, contact equivalence is based on a coupling between images and scene representations. Although contact equivalence is well known in Local Differential Topology, its use in Computer Vision is very scarce. It is implicitly embedded in some recognition approaches where one exchanges information about control points and envelopes. However, to our best knowledge, it has not been applied to multiple view geometry issues. We constraint ourselves to almost generic phenomena given by low-corank $c \leq 2$ singularities. In this way, a more stable ``geometric control'' of limit positions using envelopes of linear subspaces can be performed.

\section{Completing planar homographies}
\label{sec:planar-homographies}

This section extends conventional homographies to include the degenerate cases by considering arbitrary - including eventually singular - endomorphisms (i.e. linear maps on a vector space) acting on configurations of points. In particular, regular transformations up to scale of a 3-dimensional vector space $V^3$ belong to the group of homographies which is an open subset of the projective space $\mathbb{P}^8 = \mathbb{P}End(V^3)$. Its complementary is the set of singular endomorphisms up to scale, a cubic hypersurface defined by $det(\mathbf{X}) = 0$ for $\mathbf{X} = (x_{ij})_{0 \leq i, j \leq 2}$ and containing the fundamental subvariety $\mathcal{F}$ and the essential manifold $\mathcal{E}$.

Planar homographies represent regular transformations between two projective planes $\mathbb{P}^2 = \mathbb{P}V^3$ of 2D views. Thus, any homography is an element of the projectivized linear group $\mathbb{P}GL(3, \mathbb{R})$, where $GL(3, \mathbb{R})$ is the general linear group acting on the projective model of each view. Given a reference for $V$, each element of $GL(3, \mathbb{R})$ can be represented by a $(3 \times 3)$ regular matrix, i.e. with non-vanishing determinant. By construction, homographies (regular transformations up to scale) can not include degenerate transformations such those appearing in fundamental or essential matrices. Then, these matrices can be considered as ``degenerate'' endomorphisms (represented by defficient rank matrices up to scale) of an abstract real 3D space $V$ with $\mathbb{P}V = \mathbb{P}^{2}$. 

Fundamental matrices $\mathbf{F} \in \mathcal{F}$ are defined by degenerate bilinear forms $\mathbf{x} \mathbf{F} \mathbf{x}' = 0$ linking pairs $(\mathbf{x}, \mathbf{x}')$ of corresponding points. The set of pairs of corresponding points is called the join of two copies of $\mathbb{P}^2$. This join is defined by the image of the Segre embedding $s_{2,2}: \mathbb{P}^{2} \times \mathbb{P}^{2} \hookrightarrow \mathbb{P}^{8}$ giving a four-dimensional variety of $\mathbb{P}^8$ that determines the 7D subvariety $\mathcal{F}$ of singular endomorphisms up to scale. Addition of the singular cases ``completes'' the homographies (regular transformations), treating fundamental and essential matrices as degenerate transformation between two projective planes inside the set of a completion or homogeneous endomorphisms.

To understand how transformations can be extended from a geometric to a kinematic framework, it is convenient to introduce the differential approach for the regular subset. In terms of algebraic transformations, one mus replace the Lie group $G$ of regular transformations by its Lie algebra $\mathfrak{g} := T_e G$ where $e$ is the neutral element of $F$ (the identity matrix for matrix groups); in particular, $T_e Aut(V) = End(V)$. As usual in Lie theory, $\mathbf{A}, \mathbf{B}, \mathbf{C} \ldots $ denote the elements of the group $G$, and $\mathbf{X}, \mathbf{Y}, \mathbf{Z}, \ldots$ the elements of its Lie algebra $\mathfrak{g} := T_e G$. In particular, any endomorphism $\mathbf{X} \in \mathfrak{g} \ell(3) := T_{I}GL(3)$ can be described by a matrix representing a point $\mathbf{x} \in \mathbb{P}^8$ up to scale. 

The exponential map $exp: \mathfrak{g} \rightarrow G$ is a local diffeomorphism (with the logarithm as inverse) that can be applied to degenerate matrices for $n = 3$. In general, the set of homographies is an open set of $\mathbb{P}^N$ where $N = (n+1)^2 - 1$, whose complementary is given by the algebraic variety of degenerate matrices, i.e. $\mathcal{D} := \{ \mathbf{D} \in \mathbb{P}^N\ \mid \ det(\mathbf{D}) = 0\})$, where $det(\mathbf{D})$ is the determinant of $\mathbf{D}$. We are interested in a better understanding of degeneracy arguments from the analysis of pencils (in fact tangent directions) passing through a lower rank endomorphism. The ``moral'' consists of the following simple remark:  the original action given by a matrix product, induces an action on linear $(k+1)-dimensional$ subspaces by means the $(k+1)$-exterior power of the original action. Next paragraph illustrates this idea with a simple example.

In particular, a line $L$ represents a pencil (uni-parameter family) of endomorphisms $\{\mathbf{H}_{\lambda}\}_{\lambda \in \mathbb{P}^1}$, i.e. a linear trajectory in the ambient space $\mathbb{P}^N$ where $N = (n+1)^2 - 1$. A general line has $n+1$ degenerate endomorphisms corresponding to the intersection $L \cap \mathcal{D}$ denoted by $\mathbf{D}_1, \mathbf{D}_2, \ldots, \mathbf{D}_{n+1} \in \mathcal{D}$. Inversely, the generic element of the linear pencil $\mu_i \mathbf{D}_i + \mu_j \mathbf{D}_j$ for $1 \leq i < j \leq n+1$ is a homography away from the variety $\mathcal{D}$ of degenerate endomorphisms\footnote{This justifies perturbation arguments or, alternately, it prevents against the indiscriminate use of linear interpolation for a non-linear variety as the cubic hypersurface $\mathcal{D}$.}.

For $n = 2$, the intersection $L \cap \mathcal{D}$ of a general projective line $L \subset \mathbb{P}^{8}$ with the cubic algebraic variety $\mathcal{D}$ defined by $det(\mathbf{D}) = 0$ gives generically three different degenerate endomorphisms. In particular, if $L$ is tangent to $\mathcal{D}$ at least two elements of $L \cap \mathcal{D}$ can coalesce. An ordinary tangency condition is represented by $2 \mathbf{d} + \mathbf{d}'$, where $2 \mathbf{d}$ (resp. $\mathbf{d}'$) represents a tangency (resp. simple) contact point corresponding to the intersection of $L$ with $\mathcal{D}$. Linear pencils of matrices representing endomorphisms are interpreted as secant lines in the projective ambient space.


In general, $k$-secant varieties $Sec(k; X)$ to a variety $X \subset \mathbb{A}^{n}$ are defined by the set of points lying in the closure of $k$-dimensional subspaces $L^{k}$ generated by $(k+1)$-tuples of affine independent points generating $k$ linearly independent vectors. They can be formally constructed by using the $k$-th exterior power $\wedge^kV$ of the underlying vector space $V$ that allows to manage $k+1$-tuples of points for $k \geq 1$. This statement can be adapted to the underlying vector space of the Lie algebra $\mathfrak{g} := T_e G$ with its natural stratification by the rank of any classical group $G$. The locally symmetric structure is the key for extending the concept in the presence of singularities. Although this construction is general for $End(V)$, it can be constrained to manage eventually degenerate tensors. Actually, this approach allows to connect old based-perspective methods using homographies with tensor-based methods.

\subsection{Fundamental variety}
\label{sec:fundamental}

This subsection highlights the geometry of subvarieties parameterizing rank deficient endomorphisms (up to scale) for a three-dimensional vector space $V$.

The graph of a planar homography $\mathbf{H}$ is given by the set of pairs of corresponding points $(\mathbf{x}_i, \mathbf{x}_i') \in \mathbb{P}^{2} \times \mathbb{P}^{2}$ contained in two views modeled as projective planes fulfilling $\mathbf{H}(\mathbf{x}_i) = \mathbf{x}_i'$. From a global point of view, the ambient space is given as the image of the Segre embedding $s_{2,2}: \mathbb{P}^{2}\times \mathbb{P}^{2} \hookrightarrow \mathbb{P}^{8}$, i.e. it is a $4$-dimensional algebraic variety given at each point by the intersection of four functionally independent quadrics\cite{faugeras1993three}. As $Im(s_{2,2})$ parameterizes the set of bilinear relations between two projective planes, and each projective plane has a projective reference given by $4$ points, a general homography $\mathbf{H}$ can be described in terms of two $4$-tuples of points that can be re-interpreted as the eight (nine up to scale) projective parameters of a general matrix $\mathbf{H}$.

The space of $(3 \times 3)$-matrices $\mathbf{X}$ up to scale is a projective space $\mathbb{P}^8$, which is homogeneous by the action of the projective linear group $\mathbb{P}GL(9)$. The projective linear group $\mathbb{P}GL(3)$ induces an action on $\mathbb{P}End(V)$ that breaks the initial homogeneity of $\mathbb{P}^8$ due to the rank stratification given by three orbits. Each orbit is characterized by the rank constancy of a representative matrix. In particular, if $\mathbb{M}_r$ denotes the algebraic subvariety of matrices (up to scale) of rank $1 \leq r \leq 3$, then there is a natural rank stratification $\mathbb{M}_1 \subset \mathbb{M}_2 \subset \mathbb{M}_3$. Planar homographies may be viewed as elements of $\mathbb{M}_3 \backslash \mathbb{M}_2$ up to scale. This rank decomposition is restricted in a natural way to the fundamental variety $\mathcal{F}$.

\subsubsection{Canonical rank stratification}

This subsection includes some results regarding the endomorphisms $End_r(V)$ of a vector space $V \subset \mathbb{R}^3$, where $r$ denotes the rank of a generic element. The first result provides a description of endomorphisms $End_3(V)$ and its singular locus corresponding to degenerate endomorphisms $End_2(V)$. The second result gives its structure as a locally homogeneous space, i.e. as a disjoint union of $G$-orbits by the action of $GL(3)$ on the vector space of $\mathfrak{g}$. As usual, their elements are regular or eventually degenerate matrices, but their meaning is different as Lie group or Lie algebra, respectively.

\begin{proposition}
For any three-dimensional vector space $V$: a) the set of singular endomorphisms $End_2(V)$ is a algebraic variety of codimension $1$ given by a cubic hypersurface for $n = 2$, which is a \emph{subregular orbit} by the action of $\mathbb{P}GL(3)$ corresponding to ``subregular'' elements located in the adherence of the set of homographies in $\mathbb{P}^{8}$; b) its singular locus is given by rank $1$ \emph{degenerate endomorphisms} $End_1(V)$, which is a codimension $4$ manifold (smooth subvariety) diffeomorphically equivalent to $\mathbb{P}^{3}$ 
\end{proposition}

\begin{proof}
a) It is proved taking into account the characterization of singular endomorphisms by the vanishing of the determinant of a generic $3 \times 3$ matrix. 
b) By taking the gradient field in $\mathbb{P}^{8}$, its singular locus is locally described by the vanishing of determinants of all $(2\times 2)$ minors of a generic matrix $\mathbf{A}$ representing any endomorphism up to scale. Using $(a_{ij})$ as local coordinates in $\mathbb{P}^{8}$, if $a_{00} \neq 0$, then a local system of independent equations (local generators for the ideal of the determinantal variety representing rank 1 matrices) is locally given by

\begin{equation}
\begin{matrix}
a_{11} - a_{01} a_{10} \\ 
a_{12} - a_{02} a_{10} \\
a_{21} - a_{01} a_{20} \\ 
a_{22} - a_{02} a_{20}
\end{matrix}
\end{equation}
in the open coordinate set $D_{+}(a_{00}) := \{ \mathbf{a} \in \mathbb{P}^{8} \mid a_{00} \neq 0\}$ of $\mathbb{P}^{8}$. They are functionally independent (i.e. its jacobian matrix has maximal rank) between them. Hence, they define a smooth variety of codimension 4 (differential map of the above equations has maximal rank), which is locally diffeomorphic to $\mathbb{P}^3$. The induced group action allows to extend the local diffeomorphism to a global diffeomorphism. In particular, it is locally parameterized by $a_{11}$, $a_{12}$, $a_{21}$, $a_{22}$ corresponding to elements in the complementary box of $a_{00}$ (obtained by eliminating the row and the column of $a_{00}$).
\end{proof} 

Formally, the involution on spaces that exchanges subindexes (fixed points for transposition) leaves invariant the first and fourth generators, and identifies the second and third generators between them. Such involution corresponds to a representation of the symmetric group, giving the local generators for the Veronese variety of double lines, which is isomorphic to the dual $(\mathbb{P}^2)^{\nu}$ counted twice.

Anyway, the rank stratification can be reformulated in homogeneous coordinates as follows:

\begin{corollary}
The action of $\mathbb{P}GL(3)$ on $\mathbb{P}End(V)$ gives an \emph{equivariant decomposition} in three orbits characterized by the rank of the representative matrix up to scale. In particular: 1) the set of rank 1 endomorphisms $End_1(V)$ (up to scale) is a 4D \emph{smooth manifold} whose projectivization is diffeomorphic to $\mathbb{P}^{3}$, which is a closed orbit by the induced action; 2) the set of rank 2 endomorphisms $End_2(V)$ (up to scale) is a 7D \emph{subregular orbit}; and 3) the set of homographies corresponding to regular endomorphisms (up to scale) $End_3(V)$ is the \emph{regular orbit}.  
\end{corollary}

The stratification of endomorphisms up to scale involving the projective model of planar views can be geometrically reinterpreted by reconstructing the variety $\mathcal{E}_2$ of degenerate endomorphisms as the secant variety $Sec(1, End_1(V))$ in $\mathbb{P}^8$ of the smooth manifold $End_1(V)$. Secant varieties are explained in Section \ref{sec:secant-varieties}.

The action of $GL(n+1)$ can be extended in a natural way to the $k$-th exterior power involving $(k+1)$-tuples of vectors and their transformations for $0 \leq k \leq n$. Thus, a locally symmetric structure is obtained for arbitrary configurations of $(k+1)$-tuples of vectors (or $k$-tuples of points). It is extended in a natural way to linear envelopes of $(k+1)$-dimensional vector subspaces or, in the homogeneous case, to $k$-dimensional projective subspaces giving linear envelopes for any geometric object contained in the ambient space.

The set of $(k+1)$-dimensional linear subspaces $W^{k+1}$ are elements of a Grassmann manifold $Grass(k+1, n+1)$; its projective version is denoted as $Gr_{k}(\mathbb{P}^{n})$. Grassmann manifolds are a natural extension of projective spaces. They also provide non-trivial ``examples'' for homogeneous spaces and their generalization to symmetric spaces or spherical varieties, jointly with superimposed universal structures (fiber bundles). They have been overlooked over the years despite the presence of the analysis based on subspaces in a lot of tasks. A brief introduction to Grassmannian manifolds and their applications is provided by \cite{ye2016schubert}.




\subsubsection{Secant varieties}
\label{sec:secant-varieties}

Homographies, fundamental or essential matrices can be viewed as PL-uniparametric families (linear pencils) of matrices that can be represented by secant lines. Similarly, secant planes would correspond to PL-biparametric families (linear nets) of matrices, and so on. Additional formalism is required for a systematic treatment of these families.

Secant varieties provide a PL-approach to any variety $X$ relative to any immersion $f: X \rightarrow \mathbb{P}^{N}$. They are given as the closure $Sec(k, X)$ of points $z\in L^{k} \subset \mathbb{P}^{N}$ where $L^{k} = <x_0, \ldots, x_k>$ with  $x_0, \ldots, x_k \in f(X)$ are linearly independent. The $k$-th \emph{secant map} associates to each collection of $k+1$ l.i. points $x_0, \ldots, x_k$ their linear span $L^{k} = <x_0, \ldots, x_k>$. The closure of the graph is called the \emph{secant incidence variety}. The projection of the last component on the Grassmann manifold $Grass_{k}(\mathbb{P}^{N}) = Grass(k+1, N+1)$ is called the $k$-th secant variety (of secant $k$-dimensional subspaces) of $X$ and it is denoted by $S_{k}(X)$ as subvariety of $G_{k}(\mathbb{P}^{N})$.

Since ordinary incidence conditions $p \in L$ are invariant by the action of the projective group, secant incidence varieties represent projectively invariant conditions too. These conditions extend the well-known tangency conditions $z \in T_{x}Y$. Next, we provide a classical definition for smooth manifolds:

\begin{definition}
For any embedding $M^{m} \hookrightarrow \mathbb{P}^{N}$ of a connected regular $m$-dimensional manifold $M$, the \emph{secant variety} $Sec(1, M)$ (also called ``chordal variety'' in the old terminology) is defined by the closure of points $z \in \mathbb{P}^{N}$ lying on lines $\overline{xy}$ (called ``chords'', also), where $(x, y) \in M \times M - \Delta_M$ are different points belonging to $M$, where $ \Delta_M  := \{ (x, y) \in M \times M \mid x = y\}$ is the diagonal of $M \times M$.
\end{definition} 

Obviously, if $2m \geq N$ the secant variety $Sec(1, M)$ fills out the ambient projective space. More generally, the following result is true: 

\begin{lemma}
If $M$ is a $m$-dimensional smooth connected manifold, the \emph{expected} dimension of $Sec(1, M)$ is equal to $min(2m+1, N)$. 
\end{lemma}

The lemma is a consequence of a computation of parameters on connected smooth varieties. The dimension of the secant variety can be lower, but exceptions are well-known for a specific type of low-dimensional varieties called Severi varieties\cite{zak1986severi}. In particular, the chordal variety of the $m$-dimensional Veronese variety has dimension $2m$, instead of the expected dimension $2m+1$, providing the first non-trivial example of a Severi variety. More generally, if $M$ is a $m$-dimensional connected manifold and $2m+1 \geq N$, as $Sec(1, M)$ is a connected variety, then $Sec(1, M) = \mathbb{P}^{N}$ (see \cite[Page~40]{shafarevich2013basic} for more details about the Veronese embedding).

An alternative description for a secant variety can be provided in a purely topological way. Let define the diagonal of the product $M \times M$ as $\Delta_M$, i.e. the set of pairs $(x, y) \in M \times M$ such that $x = y$. If $M$ is a smooth $m$-dimensional variety, then $M$ is diffeomorphic to $\Delta_M$ through the diagonal embedding. Hence, the normal bundle $\mathcal{N}_{\Delta_M}$ is isomorphic to the tangent bundle $\tau_M$. Note that $\tau_{M \times M} = \tau_M \oplus \tau_M$. This topological description is useful to detect ``regular'' directions thorugh the singular locus:

Each pair $(\mathbf{x}, \mathbf{y}) \in M \times M - \Delta_M$ can be mapped to the line $\ell = <\mathbf{x}, \mathbf{y}>$ (also denoted as  $\mathbf{x} \times \mathbf{y}$). This map defines a morphism $\sigma: M \times M - \Delta_M \rightarrow Grass_1(\mathbb{P}^N)$ called the \emph{secant map} of lines. The closure of its image contains the set of tangent lines to $M$, which correspond to limit positions of secant lines when $\mathbf{x}, \mathbf{y}$ coalesce in one point $(\mathbf{x}, \mathbf{x}) \in \Delta_M$.

The closure of the graph $\Gamma_{\sigma}$ of the secant map of lines is a $(2m+1)$-dimensional incidence smooth projective variety in the product $\mathbb{P}^N \times \mathbb{P}^N \times Grass_1(\mathbb{P}^N)$ whose projection on last component is, by definition, the $1$-secant $(2m+1)$-dimensional subvariety $S_1(M)$ of $Grass_1(\mathbb{P}^N)$. The tangent space at each point $\mathbf{x} \in M$ can be described as the set of tangent lines to curves having a contact of order 2 with $\mathbf{x}$. This definition is formally extended to the singular case by taking local derivations. However, we use a more simplistic approach based on the continuity arguments and a simple computation of parameters, which gives the following result:

\begin{proposition}
The incidence variety of secant lines contains the $2m$-dimensional space $TM$ of the tangent bundle $\tau_M$ of the manifold $M$. Its projection 

$$ S_1(M) \subset Grass_1 (\mathbb{P}^N) $$

on the image of $Sec(1, M)$ by the secant map $\sigma$ is a $(m+1)$-dimensional subvariety of the Grasmannian of lines.
\end{proposition}

These descriptions show how secant lines can be understood in terms of the geometry of the ambient projective space or, alternately, in terms of the geometry of Grassmannians of lines $G_1(\mathbb{P}^n)$. The arguments are extended to higher dimension and singular varieties in \cite{fulton1984intersection}. Furthermore, they correspond to decomposable tensors which are useful for estimation issues, also.

\subsubsection{Secant to degenerate fundamental matrices}

Results in previous subsection allow to manage degeneracies in regular transformations and to perform a PL-control in terms of limit positions of secant varieties. Its extension to singular cases requires also the following result:

\begin{proposition}
Let denote the three-dimensional algebraic variety of degenerate rank $1$ fundamental matrices as $\mathcal{F}_1^3$, then $Sec(1, \mathcal{F}_{1}^{3}) = \mathcal{F}_{2}^{7}$ and $Sec(1, \mathcal{F}_{2}^{7}) = \mathbb{P}^{8}$.
\label{prop:secant-fundamental}
\end{proposition}

\begin{proof}

It suffices to prove that if $rank(\mathbf{F}) = rank(\mathbf{F}') = 1$, then $rank(\mathbf{F} + \lambda \mathbf{F}') \leq 2$, i.e. $\mid \mathbf{F} + \lambda \mathbf{F}' \mid = 0$. So, let define $\mathbf{F} = (\mathbf{f}_1 \ \mathbf{f}_2 \ \mathbf{f}_3)$, where $\mathbf{f}_{i}$ is the $i$-th column of the matrix $\mathbf{F}$ for $1\leq i\leq 3$. Then, $\mid \mathbf{F} + \lambda \mathbf{F}' \mid$ is computed as the arithmetic sum (up to sign) of determinants which are always null. More explicitly,

\begin{align*}
\mid \mathbf{F} + \lambda \mathbf{F}' \mid & = \mid \mathbf{f}_1 \ \mathbf{f}_2 \ \mathbf{f}_3 \mid \\
+ & \lambda ( \mid \mathbf{f}_1 \ \mathbf{f}_2 \ \mathbf{f}_3' \mid + \mid \mathbf{f}_1 \ \mathbf{f}_2' \ \mathbf{f}_3 \mid + \mid \mathbf{f}_1 \ \mathbf{f}_2 \ \mathbf{f}_3 \mid )\\
+ & \lambda^2 (\mid \mathbf{f}_1 \ \mathbf{f}_2' \ \mathbf{f}_3' \mid + \mid \mathbf{f}_1' \ \mathbf{f}_2 \ \mathbf{f}_3 \mid
+ \mid \mathbf{f}_1' \ \mathbf{f}_2' \ \mathbf{f}_3 \mid ) \\
+ & \lambda^3 \mid \mathbf{f}_1' \ \mathbf{f}_2' \ \mathbf{f}_3' \mid
\end{align*}

The first and last summand vanish since $\mathbf{F}, \mathbf{F}' \in \mathcal{F}_1$. By developing each determinant by the elements of the column (Laplace) and by using that all $2 \times 2$-minors of $\mathbf{F}$ and $\mathbf{F}'$ vanish the proposition is proved.

\end{proof}

This result can be also applied to symmetric matrices up to scale in the projective space $\mathbb{P}^{5}$ of plane conics:

\begin{corollary}
Let denote the two-dimensional algebraic variety of degenerate rank $1$ symmetric matrices as $\mathbb{Q}_1^2$, then $Sec(1, \mathbb{Q}_{1}^{2}) = \mathbb{Q}_{2}^{4}$ and $Sec(1, \mathbb{Q}_{2}^{4}) = \mathbb{Q}_{3} = \mathbb{P}^{5}$.
\label{cor:degenerate-symmetric}
\end{corollary}

\begin{proof}
The second equation is trivial by connectedness properties and dimensional reasons. An intuitive proof for the first equation is obtained from Proposition \ref{prop:secant-fundamental} by the involution that exchanges subindex coordinates. Intuitively, the generic element of any pencil generated by two double lines is a pair of intersecting lines\cite{zak1986severi}.
\end{proof}

The extension of the Corollary \ref{cor:degenerate-symmetric} to the space $\mathbb{P}^9$ of quadrics in $\mathbb{P}^{3}$ is meaningful for 3D reconstruction issues. Indeed, a projective compactification of essential manifold $\mathcal{E}$ is isomorphic to a hyperplane section of the variety of quadrics in $\mathbb{P}^3$ of rank $r \leq 2$ (see below), whose elements are pairs of planes (eventually coincident). 

Nevertheless their simplicity, these results are useful because provide a general strategy to manage degeneracies. In particular, for each \emph{degenerate} fundamental matrix  $\mathbf{F} \in \mathcal{F}_1$, a generic segment $\mathbf{F} + \lambda \mathbf{F}'$ connecting two rank 1 fundamental matrices gives a generic rank 2 fundamental matrix. As consequence, a generic perturbation with any PL-path removes the indeterminacy, and recovers a generic rank 2 fundamental matrix. This perturbation method (valid for stratifications with ``good incidence properties'' for adjacent strata) provides a structural connection between fundamental matrices and homographies, which can be extended to essential matrices (see Section \ref{sec:essential-remark}).

\subsection{Removing indeterminacies}

There are different kinds of indeterminacies for multilinear approaches including structural relations between pairs (given by fundamental and essential matrices, typically) or triplets of views (trifocal tensors, e.g.). All of them can be interpreted as singularities of the variety of corresponding tensors, including the fundamental variety $\mathcal{F}$ or the essential variety $\mathcal{E}$ of $\mathbb{P}End(V^3)$. Indeterminacies can be removed using secant lines to the singular locus $\mathcal{F}_1 := Sing(\mathcal{F}_2)$ of $\mathcal{F}$ (introduced in Section \ref{sec:fundamental}) or more generally $k$-th exterior powers.

Secant lines allows to control degenerated matrices using chords cutting singular varieties. These chords interpolate between more degenerate cases to recover a valid case. This subsection explains an alternative recovery strategy based on the ``recent history'' of tangent vectors to the camera poses. This also requires a careful analysis of tangent spaces to matrices in terms of their adjoint matrices.

\subsubsection{A naive approach}
\label{sec:naive-approach}

A regular transformation can be described by a matrix $\mathbf{A}$ with non-vanishing determinant or, from a dual viewpoint, by the adjoint matrix $adj(\mathbf{A})$. Entries $A_{ij}$ of the adjoint matrix are the adjoint of each element $a_{ij} \in \mathbf{A}$ up to scale; in the regular case, one must multiply with $det(\mathbf{A})^{-1}$. Hence, entries of the adjoint matrix represent up to scale the gradient field of the array $\mathbf{a}$ linked to $\mathbf{A}$.

For non-regular matrices (i.e. matrices $A$ with vanishing determinant) both descriptions are no longer equivalent between them. Let study the simplest case for $n = 3$ where the adjoint can be extended to non-regular matrices. In the regular case, the adjoint is formally described as the second exterior power $\bigwedge^{2} \mathbf{A}$ of $\mathbf{A}$. The action of $GL(3, \mathbb{R}) = Aut(V)$ on $End(V) = \mathfrak{g}$ induces an action of the second exterior power on the space of 2-dimensional subspaces (bivectors), whose projectivization represents lines of the projective plane associated to the image plane. Thus, the Adjoint map $adj: \mathbf{A} \mapsto adj(\mathbf{A})$ replaces the study of loci characterized by $0D$ features by their dual $1D$ features supported by projective lines. Loci and enveloping hyperplanes are equivalent between them for regular matrices.

This naive approach has some implications to remove indeterminacy when $rank(\mathbf{F}) = 1)$. In order to understand them $\mathcal{F}_1$ must be replaced by an enlarged space that includes the different ways of approaching each element $\mathbf{F} \in \mathbb{M}_1$ by an ``exceptional divisor''. Each divisor is supported by a finite collection of hypersurfaces in the ambient space $\mathbb{M}_3$ representing different approaching ways to the singular locus, including elements of $\mathbb{M}_{2}$. This process is known in Algebraic Geometry as a blowing-up or $\sigma$-process \cite{shafarevich2013basic}.

\paragraph{An almost-trivial example} 

Bilinear maps on a vector space are represented by a $4$-dimensional arbitrary space coupled with an inner product $<X, Y> = tr(XY)$. A direct computation shows that an orthogonal basis with $Tr(X_i^2) \neq 0$ and $Tr(X_i X_j) = 0$ is given by

$$
\left(
\begin{array}{cc}
1 & 0 \\
0 & 0
\end{array}
\right), 
\left(
\begin{array}{cc}
0 & 0 \\
0 & 1
\end{array}
\right),
\left(
\begin{array}{cc}
0 & 1 \\
1 & 0
\end{array}
\right),
\left(
\begin{array}{cc}
0 & 1 \\
-1 & 0
\end{array}
\right)
$$

Note that $X_1, X_2$ and $X_3 + X_4$ are second order nilpotent operators and that the last generator $X_4$ provides the bilinear structural constraint for the symplectic geometry on $V^2$ \cite{libermann2012symplectic,guillemin1990symplectic}. In this case, $ad(X_1) = X_2$, $ad(X_2) = X_1$, $ad(X_3) = - X_3$, and $ad(X_4) = X_4$, with $det(X_1) = det(X_2) = 0$, $det(X_3) = -1$ and $det(X_4) = 1$. 

From a projective viewpoint, the set of bilinear maps on the projective line $\mathbb{P}^1$ can be reinterpreted in terms of the image of the Segre embedding $s_{1,1}: \mathbb{P}^{1}\times \mathbb{P}^{1} \rightarrow \mathbb{P}^{3}$ given by 

$$ ([x_1: x_2], [y_1, y_2]) \mapsto [x_1 y_1: x_1 y_2: x_2 y_1: x_2 y_2]$$

Let be $z_{ij} := x_i y_j$ then $z_{11}z_{22} - z_{12}z_{21} = 0$ defines a one-sheet hyperboloid in $\mathbb{P}^3$. Hence, the set of bilinear maps on $V^2$ (up to scale) can be modelled with such hyperboloid. The main novelty here concerns the ruling given by the group action on the projective lines $\ell_1 = \mathbb{P}<X_1, X_3>$ and $\ell_2 = \mathbb{P}<X_2, X_4>$. In this case, the adjoint map induces an involution that can be translated to a complex conjugation between the generators for each ruling. 

\subsubsection{A more formal approach}
\label{sec:formal-approach}

From a local topological viewpoint, degenerate endomorphisms can be studied in terms of limits of tangent vectors $\mathbf{X} = t_{\mathbf{A}} \gamma(t)$ to curves $\gamma(t)$ through $\mathbf{A}$. Such curves represent trajectories in the matrix space connecting ``consecutive'' poses for a camera. Regularity of generic points of such curves allows to define tangent vectors at isolated degeneracies by means secant lines.

A more intrinsic approach to tangency conditions must include the dual matrix for any $\mathbf{X} \in End(V)$. It is defined at each point by the adjoint matrix $adj(\mathbf{X})$ whose entries $X_{ij}$ are the signed determinants of complementary minors of $x_{ij}$. If $\mathbf{X}$ is a regular matrix, then $\mathbf{X}.adj(\mathbf{X})$ is a power of $det(\mathbf{X}) \neq 0$, i.e. a unit from a projective viewpoint. We are interested in extending this construction to singular endomorphisms by using intermediate exterior powers. Their closure in the corresponding projective space forms the variety of complete endomorphisms.

To begin with, a first order complete endomorphism representing a degenerate planar transformation is given, up to conjugation, by a pair of matrices $(\mathbf{X}, \mathbf{X}^{\nu})$, where $\mathbf{X}$ is an endomorphism of a 3D vector space $V$, and $\mathbf{X}^{\nu}$ represents its dual given up to scale by the adjoint matrix $adj(\mathbf{X}) = \bigwedge^2 \mathbf{X}$. The replacement of a matrix with its adjoint transforms any incidence condition (pass through a point for a conic, e.g.) into a tangency condition (dual line becomes tangent at a point, e.g.). Moreover, this exchange between projectively invariant conditions does not depend of the dimension. For regular homographies, there exists a natural duality between descriptions in terms of the original matrix $A$ and its adjoint matrix $adj(A)$, giving the natural duality between incidence and tangency conditions for smooth ``objects'' (endomorphisms, in our case). Hence, the only novelty appears linked to singular strata which can be illustrated by its application to the variety of singular fundamental matrices when $n = 3$. The dual construction is compatible with the induced action of $\mathbb{P}GL(3)$ on $\mathbb{P}End(V)$ and its restriction to $\mathcal{F}$. Let $\mathbf{X} \in \mathfrak{g} \ , \mathfrak{g} = T_e G$, then $\mathbf{X}^{\nu} \in \mathfrak{g}^*$.

For arbitrary dimension, the entries $X_{ij}$ of the adjoint matrix $adj(\mathbf{X}) = \bigwedge^{n} X = det(\mathbf{X})$ for $\mathbf{X} \in End(V^{n})$ are interpreted as determinants corresponding to the components of $\nabla det(\mathbf{X})$. The next iteration for $X_{ij}$ gives the determinants of $(n-2) \times (n-2)$-minors as generators for the ideal of $\nabla X_{ij}$. Their vanishing defines the singular locus of the variety $det(\mathbf{X}) = 0$. Symbolically, $\nabla^{2} det(\mathbf{X})$ can be formulated as a double iteration of the adjoint map generated by the vanishing of determinants of minors of size $n-2$ of $\mathbf{X}$. These determinants are the generators of $\bigwedge^{n-1} End(V^{n+1})$, which is the dual of $\bigwedge^{2} End(V^{n+1})$.

The description of the previous paragraph can be geometrically reinterpreted in terms of arbitrary codimension $k$ subspaces. In particular, the extension of the adjoint map can be algebraically interpreted as a gradient field. For any endomorphism $\mathbf{X}$ let consider the $n$-tuples

\begin{equation}
(\mathbf{X}, \bigwedge^2 \mathbf{X}, \ldots, \bigwedge^n \mathbf{X}) \ ,
\end{equation}

where $\bigwedge^k \mathbf{X}$ is the $k$-th exterior power of $\mathbf{X}$, whose entries are given by the determinants of the $k \times k$-minors of $\mathbf{X}$. It is an element of the exterior algebra $\bigwedge^{*} End(V)$ defined by the direct sum of exterior powers of $End(V)$. The iteration of the gradient field given by the determinant function $det: End(V) \rightarrow \mathbb{R}$ as $\mathbf{X} \mapsto det(\mathbf{X})$ can be interpreted as ``successive derivatives'' on the space of endomorphisms. Let us remark that traces of exterior powers are the coefficients of the charatceristic polynomila $\mid \lambda I - A\mid$;, which can be reinterpreted (in the complex case) in terms of eigenvalues. Thus, in this case all the information is computable in terms of SVD with usual interpretation for the ordered collection of eigenvalues. 

In arbitrary dimension, the generic case  corresponds to the \emph{regular orbit}, i.e. endomorphisms $X$ with $rank(X) = n+1$ (automorphisms). By iterating the construction of exterior powers, one can associate an algebraic invariant given by the multirank $rank(\bigwedge^{k+1} \mathbf{X}) = rank(\bigwedge^{n-k} \mathbf{X})$. Looking at Figure \ref{fig:3d}, the regular case for $n = 2$ (resp. $n = 3$) corresponds to bilinear forms with birank $(3, 3)$ (resp. the multirank $(4, 6, 4)$ with self-duality for the mid term), which can be reinterpreted in terms of quadratic forms. The case for \emph{non-regular orbits} is constructed recurrently: let $k = corank(X)$ be the dimension of $L^k = ker(X)$, the indeterminacy is removed by adding the complete bilinear as the linked quadratic forms on $L^{k}$.

For example, for any symmetric endomorphism $\mathbf{X}$ whose projectivization is a rank $1$ plane conic, there exists a double line $\ell^2 = 0$ whose kernel is the whole line. The reduced $1D$ kernel $\ell$ is also the support for a $0D$ conic on the line given by two different points (rank 2) or a double point (rank 1), which define two orbits labeled as $(1, 2)$ and $(1, 1)$ in Figure \ref{fig:3d}. Similarly, for a rank $1$ quadric supported by a double plane $L^2$, the kernel is the whole plane that supports an embedded complete conic $(q, q^{\nu})$ in the double plane with biranks $(3, 3)$, $(2, 1)$, $(1, 2)$ and $(1,1)$. Hence, the most degenerate orbits of complete quadrics have multiranks $(1, 3, 3)$, $(1, 2, 1)$, $(1, 1, 2)$ and $(1, 1, 1)$. If $k \geq 2$ the exterior powers represent geometrically the matrices acting on envelopes by $k$-dimensional tangent linear subspaces to ``any object'' contained in $\mathbb{P}^n$ of increasing dimension. In this case, (a) the action of $GL(n+1)$ induces a conjugation action of $\bigwedge^k GL(n+1)$ on itself; (b) the $k$-th exterior powers of $\mathbf{X}$ can be considered as elements of the $k$-th exterior power of $\mathfrak{g}^*$.

Essential manifold $\mathcal{E}^5$ of regular essential matrices is embedded in $\mathbb{P}(End(V^4))$, a $15$-dimensional projective space. The extension of complete homographies on $\mathbb{P}^2$ to $\mathbb{P}^3$ creates complete collineations $(\mathbf{X}, \bigwedge^2 \mathbf{X}, \bigwedge^3 \mathbf{X})$. The third component $\bigwedge^3 \mathbf{X}$ is in fact the adjoint matrix of $\mathbf{X}$. This construction provides a general framework to obtain compactifications (as complete varieties) of orbifolds corresponding to $\mathcal{F}$ and $\mathcal{E}$ as degenerate endomorphisms.

The basic idea for extending regular to singular cases is based on adding infinitesimal information from successive adjoint maps, which is interpreted as the iteration of the gradient operator applied to the determinant of square submatrices. For generic singularities (i.e. for corank $c = 1$) it suffices to replace the original formulation by its dual, which gives the tangent vector for small displacements. For singularities with corank $c \geq 2$ successive exterior powers and complete endomorphisms must be considered. This differential description allows to interpret complete endomorphisms in terms of the ``recent history'' along the trajectory. In order to simplify the developments we consider the particular case $c = 2, n = 3$.


\subsubsection{A symbolic representation}

\begin{figure}
\includegraphics[width=0.5\textwidth]{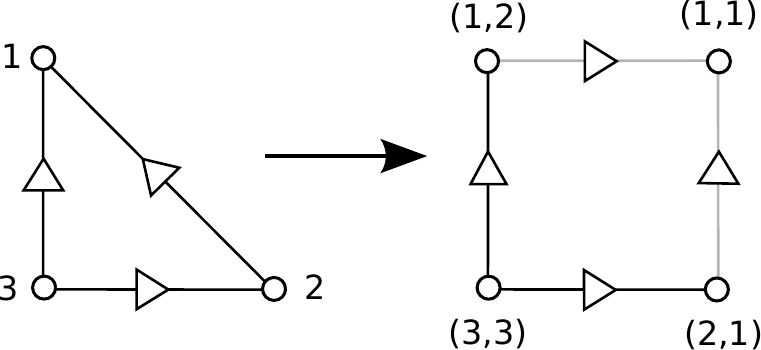}
\caption{Oriented graphs for the adjacency relations between the closures of orbits for the rank stratification of $End(V)$. Nodes represent a rank stratification whilst edges represent degeneracies in an orbit.}
\label{fig:2d}
\end{figure}

Adjacency relations between closures of orbits for rank stratification of $End(V)$ can be symbolically represented by the oriented graph of Figure \ref{fig:2d}. The vertices of the triangle represent an orbit labeled with $3, 2, 1$, according to the rank of the matrix representing the endomorphism. Oriented edges are denoted as $e_{32}$, $e_{21}$ and $e_{31}$ and represent the following degeneracies:


\begin{itemize}
    \item $3 \rightarrow 2$, corresponding to degeneracies of endomorphisms to fundamental matrices.
    \item $2 \rightarrow 1$, corresponding to degeneracies from fundamental to degenerate fundamental matrices.
    \item $3 \rightarrow 1$, corresponding to degeneracies from rank $3$ endomorphisms to degenerate fundamental matrices 
\end{itemize}

Right-side vertices of Figure \ref{fig:2d} represent the biranks corresponding to the original matrix and its generalized adjoint. A simple example is useful to illustrate the idea. Let suppose $\mathbf{X} \in End(V)$ as a diagonalizable matrix and $\lambda_1, \lambda_2$ and $\lambda_3$ as the eigenvalues of $\mathbf{A}$, then fundamental matrices have at least one vanishing eigenvalue. In the complex case canonical diagonal forms would be equivalent to diagonal matrices $(\lambda_1, \lambda_2, \lambda_3)$ whose adjoint would be $(\lambda_2\lambda_3, \lambda_1 \lambda_3, \lambda_1 \lambda_2)$, $(\lambda_2\lambda_3, 0, 0)$, and $(0, 0, 0)$, respectively. The adjoint of the third type is identically null so, initially it does not provide additional information about the most degenerate case. Information about regular elements in the degenerate support corresponding to neighbor tangent directions (infinitesimal neighborhood) must be added to avoid the indeterminacy. $\mathbb{M}_1$ must be replaced by the exceptional divisor $E$ of the blowing-up of $\mathbb{P}^8$ with center the smooth manifold $\mathbb{M}_1$. $E$ has the orbits by the adjoint action as components, represented by the pairs $(1, 2)$ and $(1, 1)$.

The blow-up of the graph $\mathcal{G}$ replaces the oriented triangle whose vertices are labeled as $3, 2$ and $1$ with a new graph $\mathcal{G}^{(1)}$. This graph is an oriented square whose vertices are labeled as $(3, 3), (2, 1)$, $(1, 2)$ and $(1, 1)$. These labels represent the biranks of pairs of endomorphisms $(\mathbf{A}, adj(\mathbf{A}))$. Biranks are relative to the generic elements of the topological closure for the orbits of the product action $G \times \bigwedge^2 G$, where $G = Aut(V) = GL(3)$, on the space $\mathfrak{g} \times \bigwedge^{2} \mathfrak{g}$, where $\mathfrak{g} = End(V) = \mathfrak{g} \ell (3)$, acting on the graph of the adjoint map.


Hence, biranks encapsulate numerical invariants for the natural extension of the action of the Lie algebra $\mathfrak{g}\ell(3)$ corresponding to $End(V) = T_{I}Aut(V)$, using pairs of infinitesimal transformations on the original and dual spaces. The simultaneous management of complete endomorphisms $(\mathbf{X}, \mathbf{X}^{\nu})$ is more suitable from the computational viewpoint. Indeed, a simple extension of SVD methods to pairs of endomorphisms allows an estimation of the generators of Lie algebras $\mathfrak{g}$ and its dual $\mathfrak{g}^{*}$ easier than the estimation of the generators for the original Lie group $G$. 

\subsection{Complete fundamental matrices}

This section adapts constructions of endomorphisms in any dimension shown in precedent section to fundamental matrices. The notion of \emph{complete fundamental matrix} is crucial to perform a control of degenerate cases from a quasi-static approach. This means that we are not taken into account the kinematics of the camera. Nevertheless, as the adjoint $\mathbf{f}^a$ represents the gradient vector field $\nabla_{\mathbf{f}^{a}} \mathcal{F}$ at the element $\mathbf{f}$, there is a measurement of ``local variation'' around each element $\mathbf{f} \in \mathcal{F}$.

\begin{definition}
The set of \emph{complete fundamental matrices} is the closure $\overline{\mathcal{F}_2^0 \times \mathcal{F}^a_2}$ of pairs $(\mathbf{F}, \mathbf{F}^a)\in $ where $\mathbf{F} \in \mathcal{F}_2$ is a fundamental matrix of rank 2, and $\mathbf{F}^a \in \mathbb{F}^a_1$ is its adjoint matrix whose entries are given by the determinants of all $2 \times 2$-minors of $\mathbf{F}$.
\end{definition} 

Degeneracies of endomorphisms up to scale linked to camera poses can be controlled with the insertion of tangential information and the restriction of the construction to $\mathcal{F}$. This information allows to represent ``instantaneous'' directions in terms of all possible directional derivatives represented by the adjoint matrix. Thus, despite $rank(\mathbf{F}) = 2$ and consequently $rank(\mathbf{F}^a) = 1$, the adjoint matrix provides a parameterized support (in fact a variety) to represent ``directions'' along which the degeneracy occurs. However, this remark is not longer true for the most degenerate cases corresponding to rank $1$ matrices that require a set of pairs of degenerate fundamental matrices.


\subsubsection{Complete pairs of fundamental matrices}

In the absence of perspective models supporting arbitrary homographies, fundamental matrices are used to avoid indeterminacies. For consecutive camera poses, two fundamental matrices $\mathbf{F}$ (for views $V_1$ and $V_2$), and $\mathbf{F}'$ (for views $V_1'$ and $V_2'$) provide a structural relation between both views. 

The dual construction $\mathbf{X} \mapsto adj(\mathbf{X})$ corresponding to the gradient $\nabla$ at each point can be restricted just to rank 2 fundamental matrices. This provides a tangential description with information about the first order evolution of $\mathbf{F}$ according to tangential constraints. From a more practical viewpoint, tangential information can be approached by secant lines in a PL-approach that connects rank 2 fundamental matrices $\mathbf{f}, \mathbf{f}' \in \mathcal{F}$. However, this construction becomes ill-defined when $rank(\mathbf{F}) = 1$ since the adjoint map is identically null. Then, tangential directions corresponding to endomorphisms of $W = Ker(\mathbf{X})$ are added to avoid this indeterminacy. 

For regular matrices the extended approach for complete objects is compatible with the differential approach given by a smooth interpolation along a geodesic path $\gamma_{i,i+1}$ connecting consecutive complete fundamental matrices $(\mathbf{f}_{i}, \mathbf{f}^{a}_{i})$ and $(\mathbf{f}_{i+1}, \mathbf{f}^{a}_{i+1})$. These are obtained as the lift of a geodesic path to the tangent space. From a theoretical viewpoint, lifting is performed by restricting the logarithm map. In practice, secant lines provide a first order approach to geodesics. 

A more detailed study of the geometry of degenerate fundamental matrices is required to recover a well-defined limit of tangent spaces in the singular case. This study must include procedures for selecting a PL-path (supported on chords) connecting the degenerated $\mathbf{f}'$ with its neighboring generic fundamental matrices. In practice, if the sampling rate is high enough there will be no meaningful difference between pairs of matrices. This generates uncertainty about the direction to approach the tangent vector. It can be solved with a coarser sampling rate along the ``precedent story''.


\subsubsection{Incidence varieties and canonical bundle}

A local neighborhood of $\mathbf{F} \in \mathcal{F}_1$ relative to $\mathcal{F}_2$ is the unit sphere corresponding to the fiber of the punctured normal bundle of $\mathcal{F}_1$ in $\mathcal{F}_2$. Normal bundles are given by a quotient of tangent bundles. Hence, tangent bundles of $\mathcal{F}_1$ and $\mathcal{F}_2 = Sec(1, \mathcal{F}_1)$ must be computed. The latter can be described as the blow-up of $\mathcal{F}_1 \times \mathcal{F}_1$ with center in the diagonal $\Delta_{\mathcal{F}_1}$. This diagonal is isomorphic to the smooth manifold $\mathcal{F}_1$ so its tangent bundle is also isomorphic to the tangent bundle of $\mathcal{F}_1$. Hence, it suffices to compute the latter and reinterpret it in geometric terms. 

Note that $\mathcal{F}_1$ is a smooth manifold diffeomorphic to $\mathbb{P}^3$, so their tangent bundles are isomorphic. Thus, $\tau_{\mathcal{F}_1} \simeq \tau_{\mathbb{P}^3}$ can be reinterpreted in terms of incidence varieties. Simplest incidence varieties in the projective plane $\mathbb{P}^2$ are given by $I_{1,2} := \{ (\mathbf{p}, \ell) \in \mathbb{P}^{2} \times (\mathbb{P}^{2})^{\nu} \mid \mathbf{p} \in \ell \}$. 

There are two projections on $\mathbb{P}^2$ and $(\mathbb{P}^2)^{\nu}$ to interpret the incidence variety as the canonical bundle of the projective plane. This elementary construction is extended to subspaces of any dimension $k$ with the canonical bundle on the Grassmannian of $k$-dimensional subspaces. An example is the incidence variety $I_{1,8} := \{ (\mathbf{a}, \ell) \in \mathbb{P}^8 \times Gr_1(\mathbb{P}^8) \mid \mathbf{a} \in \ell \}$, where $Gr_1(\mathbb{P}^8)$ denotes the Grassmannian of lines in $\mathbb{P}^8$.

The restriction to the fundamental variety $\mathcal{F}$ of the second projection on $Gr_1(\mathbb{P}^{8})$ gives the secant variety $Sec_{1}(\mathcal{F})$ to $\mathcal{F}$ that fills out all the ambient projective space. In other words, along each point $\mathbf{F} \in \mathbb{P}^8$ pass a secant line to $\mathcal{F}$. The same is also true for $E$ instead of $F$. Furthermore, an interesting result can be formulated using the same notation:

\begin{proposition}
The $7D$ secant variety $Sec_1(\mathcal{F}_1)$ is a generically triple covering of $\mathcal{F}_2$ that ramifies along the $6$-dimensional tangent variety $T_1(\mathcal{F}_1)$ (total space of the tangent bundle) corresponding to tangent lines having a double contact at each element $\mathbf{F} \in \mathcal{F}_2$.
\end{proposition}

\subsubsection{Triplets of fundamental matrices}

This Section is intended to provide an algebraic visualization of neighboring matrices at the most degenerate case, which avoids the indeterminacy at rank $1$ elements. We use a geometric interpretation of the blowing-up process explained in \cite{shafarevich2013basic} in terms of secant varieties.

In our case, the blowing-up of the variety $\mathcal{F}_1^3$ replaces each rank $1$ degenerate fundamental matrix $\mathbf{F} \in \mathcal{F}_1^{3}$ with a 3D subspace generated by four linearly independent vectors. They can be interpreted in terms of secant lines connecting the point $\mathbf{f}$ for matrix $\mathbf{F}$ with four independent points belonging to the subregular orbit $\mathcal{F}_2 \backslash \mathcal{F}_1$. Hence, each extended face of a generic ``tetrahedral configuration'' (see Figure \ref{fig:3d}) represents a 3D secant projective space to $\mathcal{F}$ displaying degeneracies at each $\mathbf{F}_1 \in \mathcal{F}_1$. 

The cubic hypersurface representing fundamental matrices in $\mathbb{P}^8$ is not a ruled variety. Thus, a generic triplet of fundamental matrices generates a $2$-dimensional secant plane to the hypersurface $det(\mathbf{F}) = 0$. Obviously, the variety of secant lines to $\mathcal{F}_2$ and trisecant 2-planes to $\mathcal{F}_1$ fills out the projective space $\mathcal{F}_3 = \mathbb{P}^8$ of endomorphisms up to scale. More specifically, any homography can be expressed as a linear combination of three generic fundamental matrices (similarly for essential matrices).

The nearest $3$-secant $2$-plane $<\mathbf{F}_1, \mathbf{F}_2, \mathbf{F}_3>$ for each rank $1$ fundamental matrix $\mathbf{F} \in \mathcal{F}_1$ can be computed using a metric on the Grassmannian of $2$-planes. The most common metric is the inner product $$<\mathbb{X}, \mathbb{Y}> = tr(\mathbb{X}. \mathbb{Y})$$ of $\mathfrak{g} = End(V)$. In particular, at each degenerate endomorphism $\mathbf{X}$ corresponding to a fundamental matrix $\mathbf{F}$ (rep. essential matrix $\mathbf{E}$), two different eigenvalues $\lambda_{i1}, \lambda_{i2}$ (resp. a non-zero double eigenvalue $\lambda_i$) are obtained for $1 \leq i \leq 3$.

In practice, the direction to choose as ``escape path'' should correspond to the nearest $\mathbf{F}_i$ with maximal distance in the plane ($\lambda_{1}, \lambda_{2}$) of non-null eigenvalues. This distance is determined w.r.t. the eigenvalues $(\lambda, 0)$ or $(0, \lambda)$ of the degenerate fundamental matrix $\mathbf{F}$ (similarly for the essential matrix). The application of this theoretical remark would must allow to escape from degenerate situations to avoid collisions against planar surfaces (corresponding to walls, floor, ground, e.g.) whose elements do not impose linearly independent conditions to determine $\mathbf{F}$. From a more practical viewpoint, the problem is the design of a control device able of identifying the ``best'' escape path in a continuous way, i.e. without applying switching procedures. In the next paragraph we give some insight about this issue.


\section{Extending the algebraic approach}
\label{sec:algebraic-approach}

This section explains how complete matrices, which can be read in terms of successive envelopes, provide specific control mechanisms to avoid degeneracies appearing in rank $1$ fundamental or essential matrices. 

A basic strategy to analyze and solve the indeterminacy locus of an endomorphism consists in augmenting the original endomorphism by their successive exterior powers. The properties of the adjoint matrix provide a geometric interpretation in terms of successive generalized secant envelopes by $k$-dimensional subspaces\footnote{This is valid for tangent subspaces as limits of secant subspaces in the Grassmannian}. Consecutive iteration on the adjoints can be viewed as a Taylor development so that when $rank(\mathbf{X})$ decreases, all complete objects contained in $ker(\mathbf{X})$ can be added to remove degeneracies. 

The lifting of the action of $Aut(V)$ on $End(V)$ to their $k$-th exterior powers delivers a structure as locally symmetric spaces for the set of complete objects linked to $End(V)$. Here $Aut(V)$ can be replaced with a group $G$, the dual of $End(V)$ (corresponding to take adjoint matrices) with $\mathfrak{g}^{*}$ and the induced action by the adjoint map $ad: G \rightarrow \mathfrak{g}^{*}$ giving the adjoint action $ad: G \times \mathfrak{g}^{*} \rightarrow \mathfrak{g}^{*}$. Then, the $k$-th exterior power $\bigwedge^{k} ad: \bigwedge^{k} G \rightarrow \bigwedge^{k}\mathfrak{g}^{*}$ is the natural extension of the adjoint map. This map induces the corresponding $k$-th adjoint action of $\bigwedge^{k} G$ on $\bigwedge^{k} \mathfrak{g}^{*}$ that extends the original action of $Aut(V)$ on $End(V)^{*}$. This simple construction is applicable for all actions of classical groups to remove their possible indeterminacies on Lie algebras.

The simplest non-trivial example is the pair $(k, n) = (2, 4)$, which defines a space of dimension $6 = \binom{4}{2}$ for $\bigwedge^2 V^4$. In this case, the $6 \times 6$ regular matrices of $\bigwedge^2 G$ (automorphisms of $\bigwedge^2 V^4$) act on the $6 \times 6$ arbitrary matrices of $\bigwedge^k \mathfrak{g}^*$ (endomorphisms of $\bigwedge^2 V^4$). In practice it requires to compute the determinants of $2 \times 2$-minors of a $4 \times 4$ regular matrix acting on $2 \times 2$-minors of a $4 \times 4$ arbitrary matrix. This explains the jump from original ranks $(4, 3, 2)$ of arbitrary $4 \times 4$-matrices and their second powers to ranks $(6, 3, 1)$. 

An adaptation of the general linear approach to the euclidean case provides a decomposition of the $6 \times 6$-matrices in $3 \times 3$-blocks that can be reinterpreted in terms of ordinary rotations. In addition, this construction can be adapted to bilateral (product or contact) actions in terms of double conjugacy classes. We are interested in the locally symmetric structure of the rank-stratified set of projection matrices linking scene and views models. A simple description of this structure for any space allows to propagate control strategies by using local symmetries, without using differential methods, no longer valid in the presence of singularities.



The completion of planar homographies in $\mathbb{P}^8$ to include fundamental matrices can be extended to any dimension and re-interpreted in matrix terms. To achieve this goal, it is required to consider the projectivization $\mathbb{P}(End(V^{n+1}))$ of a $(n+1)$-dimensional space $V$, to construct the rank stratification of matrices in $End(V)$ and to take the $(k+1)$-th exterior power (up to scale) of them for $0 \leq k\leq n$. The locally symmetric structure for the resulting completed space is obtained by the induced action of $GL(n+1) = Aut(V^{n+1})$ on the $(k+1)$-th exterior power $\bigwedge^{k+1}V$ of $V$. This structure justifies positional arguments for minimal collections of corresponding elements (points, lines, or more generally, linear subspaces) and controls their possible degeneracies in terms of adjacent orbits.

The construction of the above completion poses some challenges. For example, its topological description in $\mathbb{P}^N$ for $N = (n+1)^{2}- 1$ requires additional $(k+1)$-dimensional linear subspaces in the ambient projective space $\mathbb{P}^n$. Using contact constraints for linear subspaces has an equivalence to scene objects in terms of PL-envelopes by successive higher dimensional linear subspaces $L^{k+1}$. Due to space limitations, we constrain ourselves to the case $n = 3$ and the completion of essential matrices.

The fact that $End(V)$ (including degeneracies) is the Lie algebra of $Aut(V)$ (only regular transformations) links algebraic with differential aspects. Hence, exponential and logarithm maps provide a natural relation between both of them. However, as fundamental and essential matrices play a similar role for affine and euclidean frameworks (as non-degenerate bilinear relations), a common framework where both interpretations are compatible is required for an unified treatment of degenerated cases.



\subsection{Essential manifold}
\label{sec:essential-manifold}

The essential constraint for pairs $(\mathbf{p}, \mathbf{p}')$ of corresponding points from a calibrated camera is given by $^{T} \mathbf{p} \mathbf{E} \mathbf{p} = 0$. The set of essential matrices $\mathcal{E}$ is globally characterized as a determinantal variety in \cite[Section~2.2]{kileel2017algebraic}. The author explores the structure of $\mathbf{E}$ as a locally symmetric variety and a completion (not necessarily unique) obtained using elementary properties of adjoint matrices.

Any ordinary essential matrix $\mathbf{E}$ has a decomposition $\mathbf{E} = \mathbf{R}\mathbf{S}$, where $\mathbf{R}$ is a rotation matrix and $\mathbf{S}$ is the skew-symmetric matrix of a translation vector $\mathbf{t}$. Essential and fundamental matrices are related through $\mathbf{E} = \mathbf{M}_r^{T} \mathbf{F} \mathbf{M}_{\ell}$ where $\mathbf{M}_r$ (resp. $\mathbf{M}_{\ell}$) is an affine transformation acting on right (resp. left) on the source (resp. target space). In algebraic terms, they belong to the same double conjugacy class by the diagonal of the $A$-action of two copies of the affine group, where $A = R \times L$ is the direct product of right $R$ and left $L$ actions. Hence, essential matrices can be considered as equivalence classes of fundamental matrices. The following result gives a synthesis of the above considerations: 

\begin{proposition}
The variety $\mathcal{E}_i$ of extended essential matrices of rank $\leq i$ is a quotient of the variety $\mathcal{F}_i$ of extended fundamental matrices of rank $\leq i$. More generally, the stratified map $\mathcal{F} \rightarrow \mathcal{E}$ is an equivariant fibration between stratified analytic varieties for natural rank stratifications in $\mathbb{P}End(V)$.
\end{proposition}

The exchange between projective, affine and euclidean information and the analysis of degenerate situations requires a general framework where rank transitions can be controlled in simple terms. To accomplish this goal, a locally symmetric framework that represents degenerate cases must be developed. 

Degeneracies can be studied in the space of multilinear relations between corresponding points, from which the essential matrix is estimated \cite{kim2010degeneracy}. However, these estimations are performed in the space of configurations of points without considering the degeneracies of the matrices in the space of endomorphisms arising from an algebraic viewpoint. The secant line connecting two degenerate endomorphisms is translated in $8+8+1 = 17$ linear parameters in the same way as in \cite{kim2010degeneracy}. Additional constraints relative to fundamental or essential matrices help to reduce the number of parameters.

If $X$ is a variety with singular locus $Sing(X)$, the regular locus is denoted as $X^0 := X \backslash Sing(X)$. In particular, the set of regular fundamental matrices is denoted by $\mathcal{F}^0 = \mathcal{F}_2 \backslash \mathcal{F}_1$ and the set of essential matrices as $\mathcal{E}^0 = \mathcal{E}_2\backslash \mathcal{E}_1$. Rank stratification of $End(V)$ is $\mathcal{F}_1 \subset \mathcal{F}_2$ (resp. $\mathcal{E}_1 \subset \mathcal{E}_{2}$), where the subindex $k$ denotes the algebraic subvariety of matrices with rank $\leq k$, up to scale \footnote{Fundamental matrices are considered in Section \ref{sec:fundamental}.}. 

Two global algebraic and differential results to consider are the following ones: 

\begin{itemize}
\item The essential variety $\mathcal{E}$ is a $5$-dimensional degree $10$ subvariety of $\mathbb{P}^8$, which is isomorphic to a hyperplane section of the variety $V_2$ of complex symmetric matrices of rank $\leq 2$\cite{floystad2018chow}. The singular locus is isomorphic to $\mathbb{P}^3$ via the degree $2$ Veronese embedding $v_{2,3}$ whose image is the variety $V_1$ of double planes. In particular $Sec(1, V_1) = V_2$.  
\item If $T SO(3) \simeq SO(3) \times \mathbb{R}^{3} \simeq SO(3) \times \mathfrak{so}(3)$ denotes the total space of the tangent bundle of $SO(3)$, then the essential regular manifold $\mathcal{E}^0$ is isomorphic to the total space of unit tangent bundle of $SO(3)$ given by $SO(3) \times \mathbb{S}^2$. 
\end{itemize}

Nevertheless their local description in terms of Lie algebras, all the above isomorphisms are global since any Lie group $G$ is a parallelizable manifold. If $G$ is connected, the isomorphisms are infinitesimally given by the translation $T_A G = A.\mathfrak{g}$. In the euclidean framework, this description allows to decouple rotations and translations. The rest of this Section explains local and global properties of extended $\mathcal{E}$ and their relations with fundamental matrices.

\subsubsection{Parameterizing the essential manifold}
\label{sec:essential-manifold-parameter}

An essential matrix $\mathbf{E} \in \mathcal{E}^0$ is a $2$-rank matrix with two equal eigenvalues and a diagonal form $(\lambda, \lambda, 0)$, which in the complex case is projectively equivalent (up to scale) to $(1, 1, 0)$. \ac{SVD} decomposes $E$ in a product $U \Sigma V^{T}$ where $U, V$ are orthogonal and $det(U) = \underline{+} 1 = det(V)$. The sign of the determinant can be chosen without modifying the \ac{SVD}. Then, the fibration $\Phi: SO(3) \times SO(3) \rightarrow \mathcal{E}$ given by $\Phi(U, V) = U \Sigma V^{T}$ is a submersion with a $1$-dimensional kernel representing the ambiguity for choosing the basis of the space generated by the first two columns of $U$ and $V$.

The description of $\mathcal{E}$ in terms of the fibration $\Phi$ allows to decompose any element $\mathbf{E} \in \mathcal{E}$ in ``horizontal'' and ``vertical'' components for the tangent space to the product $SO(3) \times SO(3)$. This is crucial to reinterpret the decomposition in locally symmetric terms and to bound errors linked to large baselines\cite{subbarao2008robust}.

\subsubsection{A differential approach}

The set $\mathcal{E}^0$ of regular essential matrices is an open manifold that can be globally described in terms of the unit tangent bundle $\tau_{u} SO(3)$ to the special orthogonal group $SO(3)$ representing spatial rotations\cite{ma2001optimization}. Also, $\mathfrak{o}(3) := T_{I} SO(3)$ is the Lie algebra of $SO(3)$, i.e. the vector space of skew-symmetric $3 \times 3$-matrices that can be interpreted as translations in the tangent plane $T_{A} SO(3) = A. \mathfrak{so}(3)$ at each $A \in SO(3)$. Then, $\mathcal{E}^0$ is a $5$-dimensional algebraic variety contained in the total space $T_u SO(3) \simeq SO(3) \times \mathbb{S}^2$ of the unit tangent bundle $\tau_u SO(3)$. In this bundle each fiber takes only unit tangent vectors so we restrict to unit vectors $X \in \mathfrak{s}\mathfrak{o}(3)$.

The isomorphism $\mathbb{S}^{2} = SO(3)/SO(2)$ enables a reinterpretation of the unit vectors as spatial rotations modulo planar rotations. Each element determines a unique rotation axis, where the rotation through angles $\theta + \pi$ and $\theta - \pi$ are identical. Hence, $SO(3)$ is homeomorphic to $\mathbb{R} \mathbb{P}^{3}$, which provides a general framework for a projective interpretation in terms of space lines (see \cite[Section~3.2]{ma2001optimization}). Inversely, euclidean reduction of projective information can be viewed as a group reduction to fix the absolute quadric that plays the role of non-degenerate metric (see Section \ref{sec:essential-homogeneus}).

\subsubsection{A local homogeneous description}
\label{sec:essential-homogeneus}

The projective ambiguity of projective lines as rotation axis gives two possible solutions for corresponding elements of $SO(3)$ \cite[Section~3]{dubbelman2012manifold}. This ambiguity can be modelled as a reflection that exchanges the current phase with the opposite phase between them. Hence, pairs of regular essential matrices in $SO(3) \times SO(3)$, where the the second component is generated by the logarithmic map, generate a quadruple ambiguity corresponding to two simultaneous reflections. 

Using the topological equivalence between $SO(3)$ and $\mathbb{R}\mathbb{P}^3$, the ambiguity can be represented by the product of two copies of the tangent bundle of the projective space where elements $(\mathbf{x}, \mathbf{v})$ and $(- \mathbf{x}, -\mathbf{v})$ are identified by the antipodal map. This natural identification of the tangent bundle $\tau \mathbb{R}\mathbb{P}^3$ has not a kinematic meaning from the viewpoint of the ``recent history''. Hence, in order to solve the ambiguity, a $C^{1}$-constraint must be inserted into the essential matrices for precedent camera poses. However, this constraint is only valid under non-degeneracy conditions for essential matrices, i.e. the eigenvalue $\lambda$ must be non-null. Otherwise, essential matrix ``vanishes'' and it cannot be recovered.

\subsubsection{A remark for matrix representation}
\label{sec:essential-remark}

The manifold $\mathcal{E}^0$ of regular essential matrices can be visualized as a smooth submanifold of spatial homographies $\mathbb{P}GL(4)$ given as the open set of regular transformations as points of $\mathbb{P}^{15}$. These regular transformations are described by automorphisms $\mathbf{A} \in Aut(V^{4})$ of a $4$-dimensional vector space $V$ whose Lie algebra is given by $End(V^4)$, including possible degeneracies. Our aim is to study these degenerate cases by using locally symmetric properties extending $T_{R}SO(3) = R \mathfrak{so}(3)$. Metric distortions must be avoided when approaching to the singular locus.

The vector space $End(V^4)$ of $4 \times 4$-matrices $X$ has a natural stratification by the rank $rank(X)$ denoted by $W_1 \subset W_2 \subset W_3 \subset W_4$. Here, $W_i$ represents the algebraic variety defined by the vanishing of all determinants of size $(i+1) \times (i+1)$, which are endomorphisms of rank $\leq i$. More concretely, ordinary homographies $\mathbf{A} \in \mathbb{P}GL(4)$ are represented by points of the open set $W_4 \backslash W_3$. Obviously, $W_3$ is the natural generalization of fundamental matrices given by $det(\mathbf{X}) = 0$. 

In this framework, the variety of secant lines $Sec_{1}(\mathcal{E})$ to $\mathcal{E} \simeq SO(3) \times \mathbb{S}^{2}$ in $\mathbb{P}^{15}$ is a $11$-dimensional projective variety. Similarly, the variety of secant planes $Sec_2(\mathcal{E})$ to $\mathcal{E}$ fills out the ambient space $\mathbb{P}^{15}$. This means that any spatial homography of $\mathbb{P}^3$ can be described by three essential matrices, but not in a unique way. Even more, secant varieties to rank-stratified varieties of projective endomorphisms can be described in terms of locally symmetric varieties. The simplest case corresponds to symmetric endomorphisms representing eventually degenerate conics or quadrics. 

\begin{figure}
\includegraphics[width=0.5\textwidth]{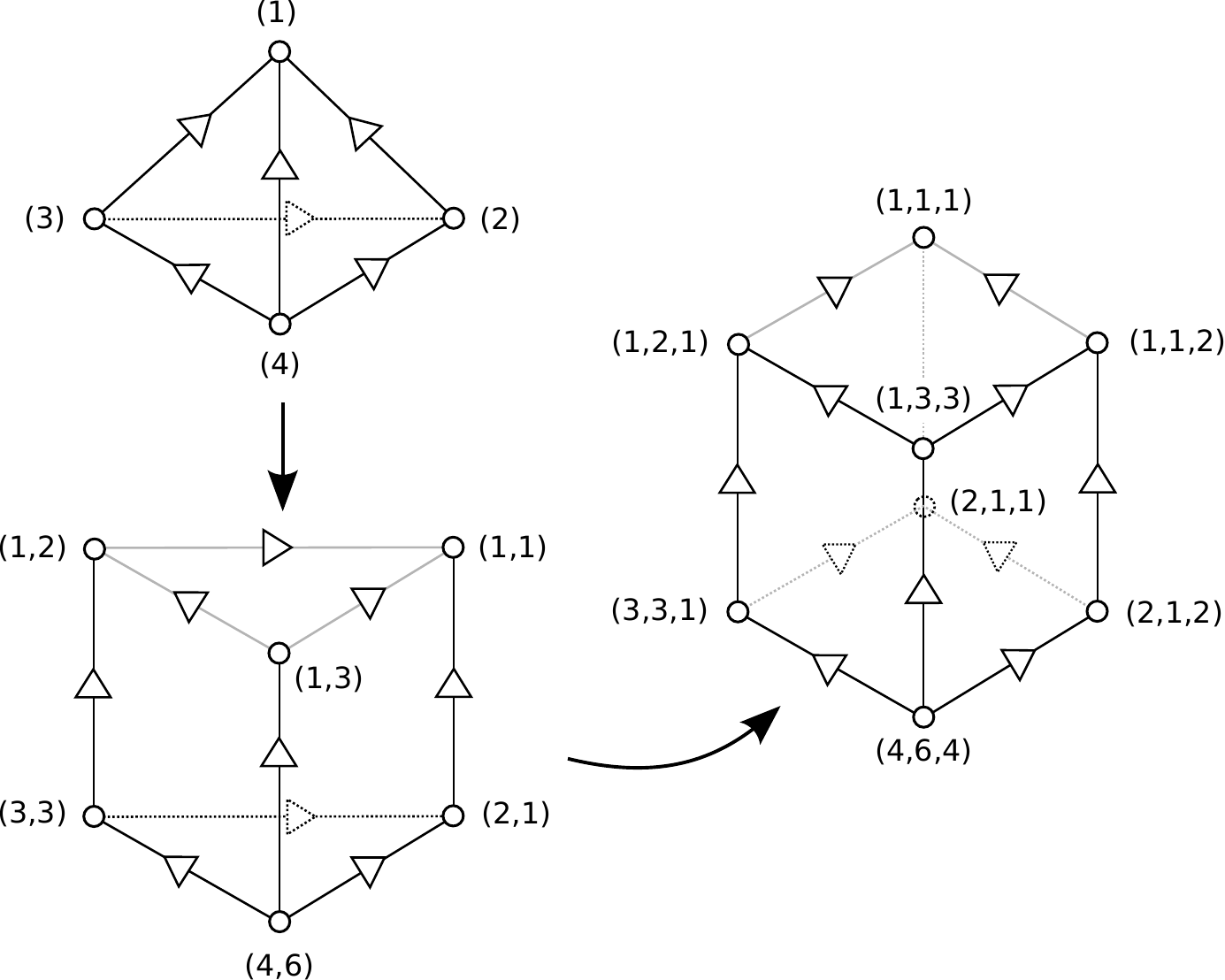}
\caption{Symbolic representation of the orbits generated by the action of $\mathbb{P}GL(4)$. Dotted edges represent hidden edges. Grey-colored edges are new edges generated after a blow-up of a vertex in a previous graph. The oriented tetrahedral graph gives the oriented triangular prism and this is finally converted into the oriented cube.}
\label{fig:3d}
\end{figure}

In particular, Figure \ref{fig:3d} illustrates how to extend Figure \ref{fig:2d} to the third dimension for the symmetric case. These orbits are induced by the action of $\mathbb{P}GL(4; \mathbb{C})$ on complete symmetric complex endomorphisms. The first blow-up in the graph replaces the vertex labeled as $1$ with the opposite face (with the same orientation). This vertex represents the endomorphisms of rank 1 up to scale or $\mathbb{E}_1$, They are completed in the new face representing the three orbits of the variety $\mathbb{E}_2$ of 1-secants $Sec(1, \mathbb{E}_{1})$. Next, the second blow-up at each minimal vertex at each height replaces such vertex by the opposite side in the height with the same orientation. The result is an oriented cubical 3D graph whose vertices represent orbits by the induced action of $GL$ on successive exterior powers. Vertices are labeled according to the multirank of each symmetric endomorphism and their respective envelopes by linear subspaces.

\subsection{The adjoint representation}
\label{sec:adjoint-representation}

Any element of a Lie group $G$ can be lifted to its Lie algebra $\mathfrak{g}$ with the log map $log: G \rightarrow \mathfrak{g}$, the local inverse of the exponential map $exp: \mathfrak{g} \rightarrow G$. The adjoint representation $G \rightarrow Aut(G)$ models $G$ as a matrix group in terms of its conjugation automorphism $\Phi_g$ defined by $\Phi_g(h) := ghg^{-1}$ for every $g \in G$. From the differential viewpoint, the adjoint representation of $G$ is computed as the differential map of $\Phi_g$ at the identity $e \in G$, i.e. $A: G \rightarrow Aut(\mathfrak{g})$ defined by $A(g) := d_e \Phi_g$.

Its dual is given by the coadjoint action $K : G \rightarrow Aut(\mathfrak{g}^{*})$ defined by the adjoint representation of the inverse element $g^{-1}$. Orbits by the coadjoint action support a symplectic structure\cite{kirillov2004lectures}. 


\subsubsection{Some meaningful examples}

There are two meaningful examples of the adjoint representation for our case:

\begin{itemize}
\item If $G = GL(3) = Aut(V)$, where $V$ is a 3D vector space, then $\mathfrak{g} = End(V)$ is naturally stratified by the rank. The coadjoint action corresponds to orbits of adjoint matrices. Its compactification has been described in Section \ref{sec:fundamental}.
\item If $G = SO(3)$, then the adjoint representation is $SO(3) \rightarrow Aut(\mathfrak{so}(3))$. It maps each spatial rotation $g \in SO(3)$ into the automorphism $\mathfrak{so}(3) \rightarrow \mathfrak{so}(3)$ representing an ``infinitesimal'' displacement between two consecutive poses as a translation in the tangent space. The adjoint action can be interpreted as the action induced by the differential of the Adjoint representation of $G$ in $Aut(G)$ given by ordinary algebraic conjugation\footnote{The adjoint starts with uppercase when it refers to a group.}.
\end{itemize}

\subsubsection{Equivariant decompositions}
\label{sec:equivariant-decomposition}

A topological equivariant stratification of a $G$-space $X$ w.r.t. a $G$-action $G \times X \rightarrow X$ is a decomposition of $X$ in a union of $G$-orbits, i.e. subsets of $X$ invariant by the action of $G$. The corresponding algebraic equivariant stratification corresponds to an algebraic action. In the presence of motion, it is interesting to construct a symplectic equivariant stratification to manage packs of solutions for structural motion equations \cite{kirwan1984cohomology}. This subsection refers only to basic aspects of algebraic equivariant stratifications.

More formally, the compactification of the coadjoint action for $GL(n)$ creates an equivariant decomposition of orbits for many general groups whose canonical forms are well-known (Jordan) \cite{bahturin2005gradings}. Furthermore, this decomposition can be restricted to the coadjoint actions for any subgroup of $G$, such as $SO(n)$ (preservation of metric properties), $SL(n)$ (volume preservation, despite shape changes), or $Sp(n)$ (preservation of motion equations). 

The above conservation laws w.r.t. the algebraic $G$-actions provide ideal theoretical structural constraints. The problem is solved by minimizing their infinitesimal variation in their Lie algebra space. The transference of information requires a more careful study of algebraic and differential relations, developed in Section \ref{sec:algebraic-differential}. For the applications concerning this paper we have constraint ourselves to only regular or subregular orbits. 

\subsection{Relating algebraic and differential approaches}
\label{sec:algebraic-differential}

This subsection explores the relations between an equivariant completion of the essential manifold and the fundamental variety in the algebraic framework given by coadjoint actions. More concretely, we provide descriptions of such completions in terms of locally symmetric spaces obtained by compactifying the original descriptions as homogeneous spaces. From a topological viewpoint, a compactification incorporates the behavior at boundaries of a topological space to preserve regularity conditions. These compactifications include additional orbits corresponding to degeneracies in the topological adherence of regular orbits. Therefore, computations for the regular case can be extended to degenerate cases too.

\subsubsection{An affine reinterpretation}
\label{sec:affine-reinterpretation}

Essential matrices have two equal non-vanishing eigenvalues and a null eigenvalue. A basic specialization principle suggests to obtain an essential matrix with a reduction to the diagonal (on the space of eigenvalues), and the addition of constraints relative to $SO(3)$ and its Lie algebra $\mathfrak{so}(3)$. By the triviality of its tangent bundle, $T_R SO(3)$ is the translation $R \mathfrak{so}(3)$ of the Lie algebra. Considering the degeneracies of fundamental and essential matrices, it is convenient to visualize how topological arguments for regular points can be extended to singularities. In particular, complete objects allow to control degeneracies in the boundary of orbits and to minimize errors in the presence of such singularities.

In affine geometry, a completion of degenerate fundamental matrices allows to recover affine reconstructions. Euclidean reconstructions (up to scale) can be achieved by preserving the absolute conic. However, euclidean information from the completion of degenerate fundamental matrices is more difficult to retrieve. A forward construction for fundamental matrices is not true because these matrices would become identically null. Thus, it is convenient to consider pairs (resp. triplets) corresponding to secant lines (resp. planes).

The description of secant spaces as locally symmetric spaces requires some considerations about products of algebraic or infinitesimal actions. The following subsections describe the topological case, the most general one, with a basic distinction between the simplest product action and the contact action, which incorporates coupling constraints.

\subsubsection{Product action}
\label{sec:product-action}

Let $Homeom(Z)$ denote the group of homeomorphisms (bijective and bicontinuous maps) of $Z$ such that for any continuous map $f: X \rightarrow Y$ between two topological spaces the $\mathcal{A}$-action is defined by the direct or Cartesian group $Homeom(X) \times Homeom(Y)$. It acts on the space of maps as $(kfg^{-1})(x) = k(f(g^{-1})x)$ for any $x \in X$ and for any $(g, k) \in Homeom(X) \times Homeom(Y)$. Obviously, $\mathcal{R}$-action on the source space $X$, and $\mathcal{L}$-action on the target space $Y$ are two particular cases of the above $\mathcal{A}$-action. 

From a topological viewpoint, it is convenient to restrict the above topological action to the differentiable case using $Diffeom(Z)$ instead of $Homeom(Z)$ for $Z = X$ or $Z = Y$. Homeomorphisms and diffeomorphisms have been used in deformation problems~\cite{miller2001group}. Since a global description of $Diffeom(Z)$ is complex and we are interested in the local aspects, only local diffeomorphisms $Diffeom_z(Z)$ are considered, i.e. diffeomorphisms preserving a point $z \in Z$.

The linearization of the differentiable $\mathcal{A}$-action gives the $A := R\times L$-action on the differentiable map $d_{x}f: T_{x}X \rightarrow T_{y}Y$ (locally represented by the Jacobian map $J_f$) at $x \in X$ with $y = f(x)$. If $p = dim(Y)$, then the $A$-action is given by $KJ_{f} H^{-1}$ for any $(H, K) \in GL(n, \mathbb{R}) \times GL(p; \mathbb{R})$. When $n = p = 3$, the action of $GL$ can be reduced to the action of the orthogonal group $O$. Its restriction to the diagonal in $O(3) \times O(3)$ provides the relation between fundamental and essential matrices.

More generally, the matrix expression of the $A$-action is given by the linearization of the double conjugacy classes for the $\mathcal{A}$-action. From the differential viewpoint, it can be written as the first order term of the $1$-jet $j^1_{e,e} \mathcal{A} = A$. This case comprises two linear actions acting on the Jacobian matrix at right and left simultaneously. In practice, the Lie algebras are preferred over the Lie groups so instead of taking the direct product of Lie groups $GL(n, \mathbb{R}) \times GL(p; \mathbb{R})$, its infinitesimal version can be chosen. It is defined by the direct product $\mathfrak{g}\ell(n) \times \mathfrak{g}\ell(p)$ of their Lie algebras, which provide a unified treatment including degenerate cases involving the restriction of endomorphisms to the corresponding Lie algebras $\mathfrak{g}$ for each classical group $G$. 

Furthermore, this approach connects directly with the extended adjoint representation from Section \ref{sec:adjoint-representation}. These constructions can be adapted to any other classical subgroup $H \subset G$, i.e. a closed subgroup preserving a non-degenerate quadratic, bilinear or multilinear form. This includes $SO(n)$, $SL(n)$, $Sp(n)$, and similarly for their Lie algebras. In all cases, a specific $G$-equivariant decomposition as union of $G$-orbits for the adjoint representation in the Lie algebra $\mathfrak{g}$ is obtained.

\subsubsection{The contact action}
\label{sec:contact-action}

In the differentiable framework, contact equivalence preserves the graph $\Gamma_f$ of any transformation $f: X \rightarrow Y$. Moreover, it introduces a natural coupling between actions on source $X$ and target $Y$ spaces for $f$, and consequently for $C^r$-equivalences acting simultaneously on $X \times Y$. A meaningful example for our purposes corresponds to linear maps of central projections of a bounded region of the image plane.

 
Topological invariants for the transformation $f: X \rightarrow X$ are linked to the fixed locus of $f$ \cite{fulton1984intersection}. This locus is the intersection product $\Gamma_f \cdot \Delta_X$, which corresponds to a weighted sum of pairs $(x, y) \in X \times X$ such that: a) $y = f(x)$, i.e. they belong to the graph $\Gamma_f$; b) $y = x$, i.e. they belong to the diagonal $\Delta_X := \{ (x, y) \in X \times X \mid x = y \}$). Therefore, transformation represented by a group action can be computed from pairs of corresponding points. 

This argument can be extended to $k$-tuples of points related by transformations acting on source and target spaces linked by the projection map. So, instead of removing 3D points generated from the pairs, which requires a posterior resolution of the ambiguity, they can be saved for a low-level interpretation in the ambient space. More formally, this ambiguity can be viewed as the dependence loci for sections of a topological fibration that takes values in the space of configurations. This idea is further explained in Section \ref{sec:spatial-homographies}.

\section{Spatial homographies and projections}
\label{sec:spatial-homographies}

This section is devoted to outline a procedure for anticipating changes in camera poses that can include degenerate cases too in order to provide some insight for the corresponding control devices in autonomous navigation including degenerate cases. The key is to exploit the locally symmetric structure of the space of projection maps arising from the double action. Our strategy considers linear actions on source and target spaces described by endomorphisms. These actions can act in a decoupled way (left-right actions) or in a coupled way (contact action). Moreover, they can be described in algebraic terms (using Lie groups) or in infinitesimal terms (using Lie algebras). The second approach allows to incorporate degeneracies in a natural way, and develop ``completion strategies'' by using exterior powers, as described in Section \ref{sec:algebraic-approach}. 

Eventual degeneracies in endomorphisms are managed by simultaneous actions on source and target spaces, which are subsets of the scene and views. Endomorphisms can be completed using exterior powers to achieve an equivariant stratification in terms of double conjugacy classes. This results in the aforementioned structure as a locally symmetric variety for the space of orbits.


\subsection{Decoupled vs coupled actions}
\label{sec:decoupled-actions}

Our strategy comprises two steps. The first step consist in decoupling source spaces (completed by a projective model for the 3D scene, e.g.) from target spaces (completed by a projective model for each view) for maps. The second step incorporates a more realistic coupling between both spaces, which is natural since each view is a projection of the scene.

In the first step there is a decoupling between left and right actions. This decoupling is firstly formulated in algebraic terms, and next extended to infinitesimal terms with complete endomorphisms to include degenerate cases. The main novelty w.r.t. precedent sections is the management of pairs of orbits involving source and target spaces. This structure can be adapted to any pair of classical groups related to the projective, affine or euclidean framework. 

The most regular algebraic transformations of the source space (a three-dimensional projective space $\mathbb{P}^3 = \mathbb{P}V^4$) are defined by the group of collineations or, more specifically, \emph{spatial homographies} $\mathbb{P}GL(4)$. Two meaningful subgroups are the affine group $\mathbb{A}_G^3 := GL(3) \ltimes \mathbb{R}^3$ (semidirect group of general linear group and the group of translations), and the euclidean group $\mathbb{E}_G^3 := SO(3) \ltimes \mathbb{R}^3$ (semidirect group of the special orthogonal group and the group of translations). 

A central projection $\mathbf{P}_i: \mathbb{P}^3 \rightarrow \mathbb{P}^2$ with center $\mathbf{C}_i$ is the conjugate of the standard projection $(I_3 \mathbf{O})$ by the left-right action denoted by $\mathcal{A} := \mathcal{R} \times \mathcal{L}$. Here $\mathcal{R} = GL(4)$ (resp. $\mathcal{L} = GL(3)$) acts on right (resp. left) by matrix multiplication up to scale. Hence, description of basic algebraic invariants for projection maps must be posed using double actions on the space of maps.

\subsubsection{The left-right equivalence}
\label{sec:left-right-equivalence}

For any pair $(\mathbf{K}, \mathbf{H})$ of regular transformations, the action on any central projection matrix $\mathbf{P}_i$ corresponding to $\mathbb{P}^3 \rightarrow \mathbb{P}^2$ is defined by $\mathbf{K} \mathbf{P}_i \mathbf{H}^{-1}$ up to scale. The double conjugacy class of the $(3 \times 4)$-matrix $\mathbf{P}_i$ for a regular central projection can be obtained by varying $(\mathbf{K}, \mathbf{H})$ in $GL(3) \times GL(4)$. 

The simplest double actions in multiple view geometry are pairs of rigid motions or affine transformations acting on source $\mathbb{P}^3$ and target $\mathbb{P}^2$ spaces for the central projection $\mathbf{P}_i$ with center $\mathbf{C}_i$

\begin{proposition}
Let define the double action by $\mathcal{R} \times \mathcal{L}$-action of pairs of diffeomorphisms acting on (germs of) maps $f: \mathbb{R}^n \rightarrow \mathbb{R}^p$ by double conjugacy, i.e. $\mathcal{A}f = \{ kfh^{-1} \mid \text{for all\ } (k, h)\in Diff(\mathbb{R}^{n}) \times Diff(\mathbb{R}^{p})\}$. Let $(h, k) \in \mathfrak{g}\ell(n) \times \mathfrak{g}\ell(p)$ be a pair of vector fields for $(K, H)$, then the tangent space to the left-right orbit $\mathcal{A}f$ is given by $k \circ f - df \circ h$.
\label{prop:left-right-equivalence}
\end{proposition}

\begin{proof}
By using the underlying topology of the set of projection matrices $[\mathbf{P}] \in \mathbb{P}^{11}$ (up to scale), a ``small perturbation'' $\mathbf{P}_\varepsilon$ of the projection map $\mathbf{P}$ around any element $\mathbf{x} = \{ (x_{ij})\mid 1\leq i\leq 3 \ , \ 1\leq j\leq j\}$ representing a $3\times 4$-matrix (up to scale) is given by 

\begin{align*}
\mathbf{P}_{\varepsilon}(\mathbf{x}) = \mathbf{P}(\mathbf{x} - \varepsilon h(\mathbf{P}(\mathbf{x})) + \varepsilon k(\mathbf{P}(\mathbf{x})) + \ldots = \\ 
\mathbf{P} + \varepsilon[k(\mathbf{P}(\mathbf{x})) - \frac{\partial \mathbf{P}}{\partial \mathbf{x}} h(\mathbf{x})] + \varepsilon^2[\ldots] + \ldots
\end{align*}

Hence, the result is proved by taking limits in $\epsilon$:

$$lim_{\varepsilon \rightarrow 0} \frac{\mathbf{P}_{\varepsilon}(\mathbf{x}) - \mathbf{P}(\mathbf{x}}{\varepsilon} = k(\mathbf{P}(\mathbf{x})) - \frac{\partial \mathbf{P}}{\partial \mathbf{x}} h (\mathbf{x})$$

\emph{Remark:} This is a particular case of the description of the tangent orbit to the $\mathcal{A}$-action for infinitesimally stable maps used in \cite[Section~1.6]{arnold1985singularities}.
\end{proof}

Additionally, local homeomorphisms arising from integrating the above vector fields can be constrained to those preserving a quadratic form (euclidean metric, or the absolute quadric in the projective version) or an invariant bilinear form (such as the symplectic form), then $GL(3) \times GL(4)$ can be replaced by the product of the corresponding affine or euclidean groups.

\begin{corollary}
The tangent space to the double conjugacy class $\mathbf{K} \mathbf{P} \mathbf{H}^{-1}$ of any projection matrix $\mathbf{P}$ is the $(3 \times 4)$-matrix $\mathbf{X} \cdot \mathbf{P} - d \mathbf{P} \cdot \mathbf{Y}$, where $\cdot$ is the ordinary product of matrices, and $(\mathbf{X}, \mathbf{Y})$ is the pair of fields $(k, h)\in \mathfrak{g}\ell(3)\times \mathfrak{g}\ell(4)$ corresponding to vector fields on $(\mathbf{K}, \mathbf{H}) \in GL(3) \times GL(4)$.
\end{corollary}

In our case $\mathbf{X} = ad(\mathbf{K}) = \nabla K$ and $\mathbf{Y} = ad(\mathbf{H}) = \nabla H$, with a slight abuse of notation. Actually, this double action is implicit in the original formulation of the \ac{KLT}-algorithm \cite{tomasi1992shape}.

\subsubsection{Incorporating the singularities}

From the topological viewpoint, spatial homographies define an open dense subset of $\mathbb{P}^{15}$ representing the ordered $(4 \times 4)$ array up to scale of entries of $\mathbf{A} \in \mathbb{P}GL(4)$. The Lie algebra of endomorphisms corresponding to homographies can also be stratified by the rank but only a small 6D submanifold of these homographies arise from a rigid motion. Thus, information from internal and external camera parameters can be recovered with the double conjugacy classes $(\mathbf{K}, \mathbf{H})$ of upper triangular matrices $\mathbf{K}$ and euclidean transforms $\mathbf{H}$.

In order to include degeneracies in the space of pairs of endomorphisms, $(\mathbf{k}, \mathbf{h})$ must display some kind of ``infinitesimal stability''. It suffices to prove that the $(3 \times 4)$-matrix $\mathbf{X} \cdot \mathbf{P} - d \mathbf{P} \cdot \mathbf{Y}$ is always regular, i.e. it has $3$ as maximal rank. In this way, it fills out the tangent space to the left-right $\mathcal{A}$-orbit of any projection matrix $\mathbf{P}$. This result is true for the regular orbit, but not necessarily for degeneracies in the Lie algebras represented by degenerate $End(\mathbb{R}^3) \times End(\mathbb{R}^4)$. However, the construction of complete endomorphisms with the exterior powers removes the indeterminacy in degeneracies of the fundamental matrix.

\subsection{Extending double conjugacy actions}

Indeterminacies in projection matrices appears when their corresponding fundamental or essential matrices are rank-deficient. In order to avoid them, the left-right equivalence for matrices (representing endomorphisms or automorphisms of vector spaces) must consider degenerate cases. In fact, this approach can be considered as a particular case of the left-right or $\mathcal{A}$-equivalence for smooth map-germs $f: \mathbb{R}^n \rightarrow \mathbb{R}^p$. More specifically, the $\mathcal{A} := \mathcal{L} \times \mathcal{R}$-action is defined by $f \mapsto k \circ f \circ h^{-1}$ for any pair of diffeomorphisms $(k, h) \in Diff_0(\mathbb{R}^n) \times Diff_0(\mathbb{R}^p)$ that preserve a point (the origin).

The linearization of the topological $\mathcal{A}$-action (pairs of diffeomorphisms) is an algebraic $A$-action that can be described in terms of pairs of automorphisms (acting on groups). From an  infinitesimal viewpoint (Lie algebra $\mathfrak{g}$), they can be seen as pairs of endomorphisms acting on the supporting vector space of $\mathfrak{g}$, to be completed if the rank is deficient. This is an immediate consequence of the description of $T_e Diff_0(\mathbb{R}^n) = GL(n; \mathbb{R})$ for the source space (similarly for the target space) of any map. Orbits are created by both actions as double conjugacy classes, but their meaning is not exactly the same: $\mathcal{A}$-action allows true deformations, whereas $A$-action only allows linear deformations, which can be reinterpreted as perspective transformations, e.g.

\subsection{Incidence varieties and multilinear constraints}
\label{sec:incidence-multilinear}

Incidence conditions between linear subspaces are given by relations, such as $L^a_i \subset L^b_j$ or $L^a_i \cap L^b_j \neq \emptyset$ where $a = dim(L^a)$ and $b = dim(L^b)$. All of them are represented by linear equations that can display deficiency rank conditions. In our case, they can also be expressed with multilinear constraints posed by the projections of the geometric objects of the scene (points and lines, mainly). Fundamental or essential matrices, trifocal tensor, or more generally multilinear tensors provide the most common examples for structural constraints between corresponding elements, such as points and lines. 

Multilinear constraints are not easy to manage due to the ambiguity in correspondences between objects and the need of efficient optimization procedures. Indeed, both problems are related since the ambiguity is solved using enough functionally independent constraints, which require optimization if the set of equations is not minimal. The simplest example is the eight-point linear algorithm that estimates the fundamental matrix\cite{hartley1997defense}. This matrix is a point $\mathbb{F}$ in the $7$-dimensional variety $\mathcal{F}$ of $\mathbb{P}^8$. A forward approach based in seven points leads to a highly non-linear procedure which is unstable and more difficult to solve. 

Optimization procedures in the space of multilinear tensors are required when redundant noisy information is available; from the algebraic viewpoing, noise can be linked to the smallest or near-zero eigenvalues. Regular tensors do not include degenerations in boundary components, because they are elements on an open set. Thus, to include degenerate tensors it is necessary to extend regular analysis which is performed in terms of complete objects (extending complete endomorphisms, for simplest tensors given by matrices). The rest of this Section remarks how eventual degeneracies can be solved by completing the information with appropriate compactifications of incidence varieties. To ease their interpretation we adopt a geometric language instead of the more formal language of \cite{thorup1988complete}.

\subsubsection{Incidence varieties}
\label{sec:incidence-varieties}

The simplest incidence variety in a projective space $\mathbb{P}^n$ fulfills $\mathbf{p} \in \ell$, where $\ell \in Grass_1(\mathbb{P}^n)$ is a line in $\mathbb{P}^n$, i.e. an element of the Grassmannian of projective lines. The set of pairs $(\mathbf{p}, \ell) \in \mathbb{P}^n \times Grass_1(\mathbb{P}^n)$ with $\mathbf{p} \in \ell$ forms the total space $E(\gamma_{1,n})$ of the canonical bundle $\gamma_{1,n}$ of $Grass_1(\mathbb{P}^n)$. A direct consequence of this construction is the following result:

\begin{proposition}
With the above notation:
\begin{itemize}
\item Epipolar constraints for corresponding points are elements of the total space $E(\gamma_{1,3})$.
\item The set of pencils $\lambda_1 \mathbf{F}_1 + \lambda_2 \mathbf{F}_2$ connecting two degenerate fundamental matrices $\mathbf{F}_1, \mathbf{F}_2 \in Sing(\mathcal{F})$ is the constraint of $E(\gamma_{1,8})$ to $\mathcal{F} \subset \mathbb{P}^8$. The topological closure of this set of pencils is the $1$-secant variety to $\mathcal{F}_1$, including cases where $\mathbf{F}_1$ and $\mathbf{F}_2$ coalesce so that the secant becomes a tangent.
\end{itemize}
\end{proposition}


The construction of the incidence variety can be extended to any kind of Grassmannians $Grass_k(\mathbb{P}^n)$, and flag manifolds denoted by $\mathcal{B}(r_1, \ldots, r_k)$ whose elements are finite collections of nested subspaces $L^{r_1} \subset L^{r_1 + r_2} \subset \ldots L^{r_1 + \ldots + r_k} = \mathbb{P}^{n}$, where $(r_1, \ldots , r_k)$ is a partition of $n+1$. If $k = 1$, then $Grass_0(\mathbb{P}^n) = \mathbb{P}^n$, and if $k = 2$, then $\mathcal{B}(k+1, n-k) = Grass_k(\mathbb{P}^n)$ with partition $(k+1, n-k)$\footnote{The notation $\mathcal{B}(r_1, \ldots, r_k)$ for flag manifold is not standard; very often $\mathcal{F}(r_1, \ldots, r_k)$ (F for flag in English) or $\mathcal{D}(r_1, \ldots, r_k)$ (D for drapeau in French) generate confusion with the space of Fundamental Matrices and the set of distributions (of vector fields, e.g.). Thus, we have chosen $\mathcal{B}$.}

In particular, the first non-trivial example of a complete flag manifold is associated to the partition $(1, 1, 1)$ of $3$. This collection represents a nested sequence of linear subspaces $V^1 \subset V^2 \subset V^3$ with projectivization $p \in \ell \subset \pi$. The stabilizer subgroup of a generic element in $\mathcal{B}(1,1,1)$ are the $3\times 3$-upper triangular matrices $\mathbb{K}$ (up to scale for projective flags) acting at left for the double conjugacy action. This formulation provides a locally symmetric structure for the left action to model changes in camera calibration. For instance, changes in focal length can be interpreted in terms of uni-parameter subgroups of the orbit in the associated flag manifold. 

These manifolds are homogeneous spaces serving as the base space of a canonical bundle that extends the properties of the simplest case of the previous paragraph. Indeed, they are ``classifying spaces'' not only for incidence, but for tangency conditions too. In general, any (eventually degenerated) collections of points and lines in the projective space can be viewed as configurations in an appropriate flag manifold $\mathcal{B}(r_1, \ldots, r_k)$. In this case, a locally symmetric structure in terms of cellular decompositions must be recovered to deal with degeneracies.

\subsubsection{Equivariant cellular decompositions}
\label{sec:cellular-decompositions}

A cellular decomposition of a $N$-dimensional variety $X$ splits the variety in a disjoint union of $k$-dimensional cells $(e_i^k, \partial e_i^k) \sim_{top} (\mathbb{B}^k, \mathbb{S}^{k-1})$ for $1 \leq k \leq N$. Here $\mathbb{S}^{k-1}$ is the $(k-1)$-dimensional sphere as the boundary of the $k$-dimensional ball $\mathbb{B}^k$. Hence, each $k$-dimensional stratum of a locally symmetric variety $X$ is a ``replication'' of a basic $k$-dimensional cell by some elementary algebraic operation (ordinary or curved reflections, e.g.). It can be represented by an oriented graph.

For example, the cellular decomposition of $\mathbb{P}^3$ is the complementary of a complete flag $p_0 \in \ell_0 \subset \pi_0 \subset \mathbb{P}^3$. They are isomorphic to affine spaces $\mathbb{A}^1 = \ell_0 \backslash p_0$ (one-dimensional cells), $\mathbb{A}^2 = \pi_0 \backslash \ell_0$ (two-dimensional cells), and $\mathbb{A}^3 = \mathbb{P}^3 \backslash \pi_0$ (three-dimensional cells), corresponding to the fixation of elements at infinity. Furthermore, such cells are invariant by the action of the affine group. A similar reasoning can be outlined using the reduction to the euclidean group by fixing the absolute quadric, the absolute conic and the circular points \cite{hartley2003multiple}.

A less trivial example is the cellular decomposition of the Grassmannian $Grass_1(\mathbb{P}^3)$ of lines $\ell \subset \mathbb{P}^3$, which is the relative localization of lines $\ell$ w.r.t. a complete flag $p_0 \in \ell_0 \subset \pi_0 \subset \mathbb{P}^3$. They are isomorphic to affine spaces given by the complementary of consecutive Schubert cycles $\sigma(a_0, a_1)$ or their dual representation $(\beta_0, \beta_1)$ with $2 \geq \beta_0 \geq \beta_1 \geq 0$. 

The extension to arbitrary Grassmmannians is a cumbersome problem (see \cite[Chapter~14]{fulton1984intersection} and \cite{griffiths2014principles}). However, their geometric meaning is simpler: they represent elements with an excedentary intersection for incidence conditions. Their dual interpretation expresses rank deficiency conditions for block matrices whose entries are subspaces, such as tangent spaces to submanifolds. In particular, for $Grass_1(\mathbb{3})$ the dual description $(\beta_0, \beta_1)$ of Schubert cycles can be easily visualized in an oriented graph with nodes located at the vertices of an ``increasing stair''. Hence, the cellular decomposition contains one 4D cell, one 3D cell, two 2D cells, and one 1D cell, which are described as follows:

\begin{itemize}
\item $(0, 0)$ does not impose conditions about lines so it represents the whole Grassmannian;
\item $(1, 0)$ imposes the constraint $\ell \cap \ell_0 \neq \emptyset$, which is a hyperplane section of the Grassmannian;
\item $(2, 0)$ is $\{ \ell \in Grass_1(\mathbb{3}) \mid \ell \subset \pi_0\}$, which imposes two conditions and it is isomorphic to the dual of $\mathbb{P}^2$
\item $(1, 1)$ is $\{ \ell \in Grass_1(\mathbb{3}) \mid p_0 \in \ell \}$, which imposes two conditions and it is isomorphic to an ordinary projective plane $\mathbb{P}^2$; 
\item $(2, 1) = \{ \ell \in Grass_1(\mathbb{3}) \mid p_0 \in \ell \subset \pi_0\}$, which imposes three conditions and it is isomorphic to a projective line $\mathbb{P}^1$;
\item $(0, 0)$ is the point representing the fixed line $\ell_0$ of the flag.
\end{itemize}

The same decomposition can be formulated in the euclidean terms on the underlying vector space $V \subset \mathbb{R}^{n+1}$ by selecting nested subspaces linked to the standard reference $\{ e_1, \ldots, e_4$ as an oriented fixed flag. Furthermore, this decomposition can be extended to any $Grass_1(\mathbb{P}^n)$ adding the vertices linked to $n-1$ steps of the stair. Two specially relevant cases for secant lines (including degeneracies) on spaces of bilinear epipolar constraints are $Grass_1(\mathbb{P}^8)$ and $Grass_1(\mathbb{P}^{15})$) that represent the ambient space for planar and volumetric collineations, respectively.

\subsubsection{Evolving fundamental matrices}
\label{sec:evolving-fundamental}

The epipolar constraint $(\mathbf{p}, \mathbf{p}') \in \mathbb{P}^2 \times \mathbb{P}^2$ can be expressed in the projective framework as $^{T} \mathbf{p} \mathbf{F} \mathbf{p}' = 0$, where $\mathbf{F}$ is the fundamental matrix. But it can also be interpreted as the simplest incidence relations $\mathbf{p} \in \ell'$ or as $\mathbf{p}' \in \ell$ in each projective plane, which are mutually dual. 

The set of pairs $(\mathbf{p}, \ell)$ fulfilling the epipolar constraint are isomorphic to the total space of the canonical bundle $\gamma_1^2$ on each projective plane $\mathbb{P}^2$. From a global viewpoint, the evolution in space and time of the epipolar constraint can be described on the tangent space $\tau_{\mathbb{P}^2}$ to the projective plane. It is well-known that

\begin{equation}
\varepsilon^{1}_{\mathbb{P}^2} \oplus \tau_{\mathbb{P}^{2}} \simeq \gamma_2^{1} \oplus \gamma_2^{1} \oplus  \gamma_2^{1} \ ,
\end{equation}

where $\varepsilon^1_{\mathbb{P}^2}$ is the trivial vector bundle on the projective plane $\mathbb{P}^2$, $\oplus$ is the Whitney sum of vector bundles, and $\gamma_2^1$ is the canonical line bundle on $\mathbb{P}^2$. This justifies the twist of a fundamental matrix for the secant line (uniparametric pencil) connecting a pair of camera locations.

\subsubsection{Evolving essential matrices}
\label{sec:evolving-essential}

In the euclidean space the epipolar constraint for corresponding points $\mathbf{p}, \mathbf{p}' \in \mathbb{E}^2 \times \mathbb{E}^2$ is represented by $^{T} \mathbf{p} \mathbf{E} \mathbf{p}' = 0$, where $\mathbf{E} \in \mathcal{E}$ is the essential matrix linked to each pair of views.

More specifically, if $\{ \mathbf{e}_i\}_{1 \leq i \leq 3}$ represents the canonical basis for $\mathbb{E}_3$, then the collection of subspaces represented by $\mathbf{e}_1$, $\mathbf{e}_1 \wedge \mathbf{e}_2$ and $\mathbf{e}_i \wedge \mathbf{e}_2 \wedge \mathbf{e}_3$ are a positive oriented flag. This flag can be viewed as the starting flag to interpret essential matrices in terms of rotations and translations.

\begin{proposition}
The local description of a essential matrix as a product of $SO(3)$ and $\mathbb{S}^2$ can be formulated globally as follows:

\begin{enumerate}
\item The essential manifold $\mathcal{E}$ is the unit sphere bundle $\mathbb{S} \tau SO(3)$ of the tangent bundle $\tau SO(3)$.
\item The evolution of the tangent bundle is $\tau (SO(3) \times \mathbb{S}^2) \simeq \mathfrak{so}(3) \oplus \tau_{\mathbb{S}^2}$. \end{enumerate}

\end{proposition}

\begin{proof}
The former is proved in \cite{subbarao2008robust}. The proof of the latter is based on the embedding $i: \mathbb{S}^2 \hookrightarrow \mathbb{R}^3$, such that there exists an isomorphism 

$$\tau_{\mathbb{S}^2} \oplus \mathcal{N}_{\mathbb{S}^2} \simeq i^{*} \tau_{\mathbb{R}^{3}} = \varepsilon^{3}_{\mathbb{S}^{2}}$$

where $\mathcal{N}_{\mathbb{S}^2} \simeq \varepsilon^{1}_{\mathbb{S}^{2}}$ is the normal bundle to the ordinary embedding $i$ of $\mathbb{S}^2$ and

$$\varepsilon^{3}_{\mathbb{S}^{2}} = \varepsilon^{1}_{\mathbb{S}^{2}} \oplus \varepsilon^{1}_{\mathbb{S}^{2}} \oplus \varepsilon^{1}_{\mathbb{S}^{2}}$$ 

is the $3$-rank trivial bundle on $\mathbb{S}^2$. Thus,

$$\tau_{\mathbb{S}^2} \oplus \varepsilon^{1}_{\mathbb{S}^{2}} \simeq \varepsilon^{3}_{\mathbb{S}^{2}}$$

However any copy of $\varepsilon^1_{\mathbb{S}^{2}}$ cannot be simplified since $\tau_{\mathbb{S}^2}$ is not topologically trivial, i.e. it is not isomorphic to the Whitney sum of two copies of $\varepsilon^1_{\mathbb{S}^2}$. The quotient by the action of $\mathbb{Z}_2$ (corresponding to the lifting of the antipodal map to the canonical bundle) gives the description of the tangent space to the projective space. This result links the information of fundamental matrices with the similar information for essential matrices.
\end{proof}







\section{Practical considerations}
\label{sec:practical-considerations}

Despite the fact that most results presented in this paper are posed in a static framework, our main motivation arise from the indeterminacies appearing at bootstrapping a mobile calibrated camera in a non-structured environment. There are many approaches in the state of the art based on perspective models working in structured or man-made scenes that take advantage of the support provided by perspective lines, vanishing points, horizon lines, e.g. However, the problem becomes more challenging in low-structured environments where a ``weighted combination'' of homographies and fundamental matrices provides a practical solution for bootstrapping. 

Generally, planar or structured scenes are better explained by a homography, whereas non-planar or unstructured scenes are better explained by a fundamental matrix. Usually, it is preferable to not assume a specific geometric model but to compute both of them in parallel \cite{mur2015orb}. Later, the best model is selected using a simple heuristic that avoids the low-parallax cases and the well-known twofold ambiguity solution arising when all points in a planar scene are closer to the camera centers \cite{longuet1986reconstruction}. In low-parallax cases both models are not well-constrained and the solution yields an initial corrupted map that should be rejected. Indeed, the quality of the tracking relies heavily in the bootstrapping of the system and, more specifically, in the choice of the most suitable geometric model.

Our heuristic approach is developed using weighted PL-paths in the secant variety $Sec(1, \mathcal{F})$ to the fundamental variety $\mathcal{F}$ that fills out the whole space $\mathbb{P}^8 = \mathbb{P}End(V)$. Each secant line $\ell \in Gr_{1}(\mathbb{P}^{8})$ cuts out $\mathcal{F}$ generically in three elements, which can coalesce in a double tangency point plus an ordinary point. The tangent hyperplane at each point of $\mathbb{F} \in \mathcal{F}$ provides the homography $\mathbb{H}$ related to the fundamental matrix $\mathbb{F}$.

Overall, the main challenge consists of retrieving a valid fundamental matrix when the estimation degenerates into a $1$-rank matrix, instead of the expected $2$-rank matrix (when the camera points towards a planar scene, e.g.). The most common approaches are based on the introduction of additional sensors, the manual specification of two keyframes in a structured scene from which the system must bootstrap \cite{klein2007parallel}, or the perturbation of the camera pose to find a close keyframe from which to recover. Each approach has its own drawbacks:

\begin{enumerate}
\item Additional sensors or devices improve the robustness and safety of the systems (a requirement for autonomous navigation), but it does not provide a scientific solution to the problem. Moreover, it is not always possible to modify the hardware of a system.
\item Manual selection of the initial keyframe pair breaks the autonomy of the system since it requires the user interaction to bootstrap. In addition, not all scenes contain a structured region to compute the homography.
\item The perturbation of the camera pose degrades the continuity in the image and scene flows since the direction of the perturbation is randomized. Also, it ignores the recent history of the trajectory, leaving the system in an inconsistent state. Proposition \ref{prop:left-right-equivalence} provides an infinitesimally stable structural result to avoid this problem.
\end{enumerate}

Our approach retrieve the recent history expressed in terms of the kinematics of the trajectory as a lifted path from the group $G$ of transformations to its tangent bundle $G \times \mathfrak{g}$, which adds the unit vector linked to the gradient to avoid the indeterminacy in the completion of $\mathbb{P}End(V)$. In presence of uncertainty due to rank deficiency, the local secant cone to the regular part of the fundamental or the essential variety can be computed. Then, the shortest chord that minimizes the angle w.r.t. the precedent trajectory is selected to avoid abrupt discontinuities which make more difficult the control.

Actually, this approach is independent of the dimension of the underlying vector space $\mathbf{V}$, and thus it could be extended to support dynamic scenes with additional structural homogeneous constraints. It is also independent of the subgroup so it can also be adapted to other classical groups, such as $SL(2)$ or $SL(3)$, which leave invariant the area or volume elements, or even the symplectic groups leaving invariant  Hamilton-Jacobi motion equations \cite{osher1988fronts}. Hence, motion analysis in dynamic environments can be performed theoretically using the same approach.

\section{Conclusions and future work}
\label{sec:conclusions}

Degeneracies in fundamental and essential matrix are a common issue for hand-held cameras traveling around a non-controlled environment. These singular matrices can be incorporated to the analysis using results from Classical Algebraic Geometry. This paper introduces the concept of complete endomorphisms to manage degeneracies, providing a geometric reinterpretation in terms of secant varieties. Instead of looking at configurations of $(k+1)$-tuples of corresponding points, our alternative approach focuses on the projective geometry of ambient spaces where tensors and projection maps live. The graph of the adjoint map for $End(V) = T\ Aut(V)$ viewed as the gradient field for matrix space provides the first example of completion. This construction is applied to the fundamental $\mathcal{F}$ and the essential $\mathcal{E}$ varieties by adding limits of tangent directions approached by secants in the ambient space $\mathbb{P}End(V)$ 

Completions of regular transformations (automorphisms in a Lie group $G$) in terms of their tangent spaces (endomorphisms in the Lie algebra $\mathfrak{g}:= T_e G$) are also extended to include the degenerate cases for projection matrices. These completions are always managed in terms of rank stratifications of spaces of matrices. Exterior algebra of the underlying vector spaces and its projectivization provides a framework to manage this rank stratification. These stratifications can be interpreted geometrically in terms of secant subspaces and their adjacent tangent subspaces of any dimension. The simultaneous completion of the transformations w.r.t. left-right action $\mathcal{A}$ on the source and target spaces or contact action $\mathcal{K}$ on the graph of the projection map $\mathbb{P}^3 \rightarrow \mathbb{P}^2$ allows a more robust feedback between image interpretation and scene reconstruction. 

The constructions of this paper admit extension concerning several topics, such as analytic presentation of limits of tangent spaces (by using appropriate compactifications, e.g.), intrinsic localization of degeneracy loci in terms of cellular decompositions (inverse image of the secant map, e.g.), preservation of locally symmetric structure of complete spaces (involving the adjoint representations), intrinsic formulation of image and scene flows (for mobile cameras, e.g.), or a relation between the motion and structure tensors (in the moment-map framework, e.g.). The results can also be adapted to the geometry of different kinds of infinitesimal transformations involving arbitrary deformations of the geometric models of views.


These developments have a direct application for bootstrapping and tracking the transformations of a camera pose in video sequences recorded in non-restricted environments with arbitrary movements. Therefore, far from being just a theoretical curiosity, the infinitesimal completion of regular transformations provides a natural and continuous framework for a unified treatment of kinematics. In particular, degenerate cases can be managed in motion prediction to increase the robustness of \acl{VO} algorithms.



\bibliographystyle{spmpsci}
\bibliography{bib.bib}

\end{document}